\documentclass[letterpaper]{article} 
\usepackage{aaai25}  
\usepackage{times}  
\usepackage{helvet}  
\usepackage{courier}  
\usepackage[hyphens]{url}  
\usepackage{graphicx} 
\usepackage{natbib}  
\usepackage{caption} 
\usepackage{algorithm}
\usepackage{algorithmic}

\usepackage{amsmath}
\usepackage{amssymb}
\usepackage{amsfonts}
\usepackage{booktabs}
\usepackage{color}
\usepackage{xcolor}
\usepackage[capitalise]{cleveref}
\usepackage{epstopdf}
\usepackage{subfigure}
\usepackage{multirow}
\usepackage{adjustbox}
\usepackage{tikz}
\usepackage{pgfplots}

\newtheorem{assumption}{Assumption}

\newtheorem{lemma}{Lemma}
\newtheorem{theorem}{Theorem}

\usepackage{newfloat}
\usepackage{listings}
\DeclareCaptionStyle{ruled}{labelfont=normalfont,labelsep=colon,strut=off} 
\floatstyle{ruled}
\newfloat{listing}{tb}{lst}{}
\floatname{listing}{Listing}
\title{Mixture of Online and Offline Experts for Non-Stationary Time Series}
\author {
Zhilin Zhao\textsuperscript{\rm 1,\rm 2,\rm 3},
Longbing Cao\textsuperscript{\rm 2},
Yuanyu Wan\textsuperscript{\rm 4}
}
\affiliations {
\textsuperscript{\rm 1} School of Computer Science and Engineering, Sun Yat-sen University, Guangzhou, China\\
\textsuperscript{\rm 2} School of Computing, Macquarie University, Sydney, Australia\\
\textsuperscript{\rm 3} Key Laboratory of Machine Intelligence and Advanced Computing, Ministry of Education, China\\
\textsuperscript{\rm 4} School of Software Technology, Zhejiang University, Ningbo, China\\
zhaozhl7@hotmail.com, longbing.cao@mq.edu.au, wanyy@zju.edu.cn
}

\usepackage{bibentry}

\begin{document}

\maketitle

\begin{abstract}
	We consider a general and realistic scenario involving non-stationary time series, consisting of several offline intervals with different distributions within a fixed offline time horizon, and an online interval that continuously receives new samples. For non-stationary time series, the data distribution in the current online interval may have appeared in previous offline intervals. We theoretically explore the feasibility of applying knowledge from offline intervals to the current online interval. To this end, we propose the Mixture of Online and Offline Experts (MOOE). MOOE learns static offline experts from offline intervals and maintains a dynamic online expert for the current online interval. It then adaptively combines the offline and online experts using a meta expert to make predictions for the samples received in the online interval. Specifically, we focus on theoretical analysis, deriving parameter convergence, regret bounds, and generalization error bounds to prove the effectiveness of the algorithm.
\end{abstract}

%

\section{Introduction}
For a non-stationary time series, the data distribution in the current time window may have appeared in the past. Therefore, can we apply the knowledge from historical data to the current time window? In this paper, we theoretically prove that this is feasible.

A common assumption in statistical learning theories for time series is that observed samples are i.i.d., or stationary in stochastic processes~\cite{TS:94}. To leverage sample dependence in non-i.i.d. processes, it is often assumed that observations come from a stationary $\phi$-mixing or $\beta$-mixing sequence~\cite{PB:10}. However, these assumptions may not hold as the distribution of real-life time series usually changes over time, making the hypothesis class not (agnostically) PAC learnable~\cite{PAC:16}. Fortunately, distribution changes in real life are often gradual, and samples in a short interval are nearly identically distributed~\cite{TS:15}. Therefore, we consider a realistic scenario involving non-stationary time series with several offline intervals of different distributions within a fixed offline time horizon and an online interval that continuously receives new samples. Once the number of received labeled samples reaches a predefined size, the online interval is converted to the last offline interval, and a new online interval begins.

Some existing methods~\cite{OP:12, DR:18} train an expert on the entire time series using off-the-shelf online optimization techniques, without considering the non-stationary nature of the data. However, the dynamic data with varying distributions can mislead the expert. Other methods~\cite{MVC:94, RM:08, PB:10} train an expert from scratch for each new online interval, which is safer but unreliable due to the scarcity of labeled samples at the early stage. Thus, it is fundamental yet highly challenging to design a learning method with tight sample complexity that outputs a hypothesis with desirable generalization. For non-stationary time series, the data distribution in the current online interval may have appeared in historical offline intervals. Therefore, a natural solution is to combine the offline experts from the offline intervals with the online expert from the current interval to address the shortcomings of the aforementioned methods.

Inspired by the mixture of experts~\cite{puigcerver2024from}, we propose Mixture of Online and Offline Experts (MOOE) to transfer knowledge from offline intervals to the online interval, addressing the non-stationarity issue. Following the paradigm of prediction with expert advice~\cite{PLG:06, MG:16}, MOOE employs a meta expert to combine the online and offline experts by adaptively weighting them according to their effectiveness. Specifically, the online expert is continuously updated in the current online interval using an existing online optimization method. Additionally, when an online interval collects enough samples to become an offline interval, all samples from this interval, along with previously obtained offline experts, are used to train the offline expert for this interval.

Theoretically, we prove that the regret of MOOE is determined by the regret of the off-the-shelf online optimization method used for the online expert. However, this can be improved if the number of maintained experts is within a bound controlled by the size of intervals and the empirical errors of the offline experts. By connecting optimization with learning theory~\cite{OL:16}, we derive the generalization error bound by jointly exploiting the regret, the properties of the loss function, the hypothesis class, and the data distribution, thereby verifying the effectiveness of our approach. Experimentally, MOOE outperforms state-of-the-art methods for handling non-stationary time series.

\section{Related Work}
\subsection{Learning Theory for non-stationary Time Series}
For non-i.i.d. processes, under the stationary and $\beta$-mixing assumptions, the early work~\cite{MVC:94} establishes the convergence rate over VC-dimension, and the work in~\cite{RM:08} presents data-dependent bounds in terms of the Rademacher complexity. By exploiting the stability properties of a specific learning algorithm, generalization bounds for $\phi$-mixing and $\beta$-mixing sequences are provided in~\cite{PB:10}. However, the mixing assumption is hard to be verified in practice. There are some attempts to relax the stationary and mixing assumptions. The uniform convergence under ergodicity sampling is shown in the work of~\cite{SES:10}. For an asymptotically stationary (mixing) process, although a generalization error is derived in~\cite{DS:13} through the regret of an online algorithm, and their analysis depends on the assumption that the output from an online learning algorithm tends to be stable, which is invalid in a dynamic environment. In~\cite{GNS:14}, the guarantee of the learning rate for nonstationary mixing processes is given by a sub-sample selection technique with the Rademacher complexity. Further, in~\cite{TS:15}, a more general scenario of nonstationary and non-mixing processes is considered, which proves the learning guarantees with the conditions of the discrepancies between distributions.

\subsection{Regret Analysis in dynamic environments}
The regret theory~\cite{RG:16} for measuring the performance has been extensively studied. The dynamic regret~\cite{DBLP:conf/icml/AnagnostidesFS23} and its restricted form~\cite{DR:15} have been introduced to manage changing environments. A basic idea behind such regrets is to compare the cumulative loss of the learned expert with several experts rather than the best one. Along this line of study, adaptive learning for dynamic regret (Ader)~\cite{DR:18} considers multiple experts with various learning rates updated by online gradient descent (OGD)~\cite{OSG:03}, and the established upper bound matches the lower bound. Another independent work for dynamic regret in a nonstationary environment is about multi-armed bandit (MAB)~\cite{DR:15}, where the work in~\cite{DMAB:19} reveals how the statistical variance of the loss distributions affect the dynamic regret bound. However, these dynamic regrets depend on the distribution changing times, which are usually unknown. When the sequence of samples is very long, the data distribution may have changed many times. As a result, the loose bound cannot measure the learned expert performance in the current interval. Another limitation is that the bound is inappropriate for analyzing experts learned on the fly because these regrets only act on observed samples.


\begin{figure*}
	\centering
	\begin{adjustbox}{width=.8\textwidth,center}
		\pgfmathdeclarefunction{gauss}{2}{%
			\pgfmathparse{1/(#2*sqrt(2*pi))*exp(-((x-#1)^2)/(2*#2^2))}%
		}
		\begin{tikzpicture}
			[
			scale = 0.5,
			node distance=1.9cm,
			MetaExpert/.style = {circle, draw=black, dash pattern={on 6pt off 3pt},align=center, anchor=east, inner sep=0, right color=green!30,left color=red!30,shading angle=-45,minimum size=0.9cm,anchor=base,font=\small},
			OfflineExpert/.style = {circle, draw=black, densely dotted,align=center, anchor=east, inner sep=0, fill=red!30,minimum size=0.9cm,anchor=base,font=\small},
			OnlineExpert/.style = {circle, draw=black, solid,align=center, anchor=east, inner sep=0, fill=green!30,minimum size=0.9cm,anchor=base,font=\small},
			Sample/.style = {rectangle, draw, solid, align=center, anchor=east, inner sep=0, fill=yellow!30,minimum size=0.4cm,anchor=base,outer sep = 0,rounded corners=0,font=\small, outer sep=0},
			ExpertSet/.style = {matrix,fill=cyan!20,draw=black,loosely dotted,align=center, outer sep = 0,inner sep = 2,rounded corners=.8ex, column sep=0.1cm, row sep=0.1cm},
			OfflineInterval/.style = {matrix,fill=red!20,draw=black,densely dotted,align=center, outer sep = 0,inner sep = 2.5,rounded corners=.8ex, column sep=0cm},
			OnlineInterval/.style = {matrix,fill=green!20,draw=black,align=center, outer sep = 0,inner sep = 2.5,rounded corners=.8ex},
			]
			
			\node[OnlineInterval] (I1) {
				\node[Sample] {$\mathbf{x}_1$};  & \node {$\cdots$} ; & \node[Sample](RS) {$\mathbf{x}_t$}; & \node {$\quad$} ; & \node {$\quad$} ;\\};
			
			\node[OfflineInterval, left of =I1,xshift = -1cm] (Ih1) {
				\node[Sample] {};  & \node {$\cdots$} ; & \node[Sample] {}; & \node {$\cdots$} ; & \node[Sample] {};\\};
			
			\node[outer sep = 0,left of =Ih1](Ih2) {$\cdots$} ;
			
			\node[OfflineInterval, left of =Ih2] (Ih3) {
				\node[Sample] {};  & \node {$\cdots$} ; & \node[Sample] {}; & \node {$\cdots$} ; & \node[Sample] {};\\};
			
			\node[OfflineInterval, below of=I1, yshift=0.7cm] (I2) {
				\node[Sample] {$\mathbf{x}_1$};  & \node {$\cdots$} ; & \node[Sample] {$\mathbf{x}_t$}; & \node {$\cdots$} ; & \node[Sample] {$\mathbf{x}_B$}; \\};
			
			\node[OnlineExpert, above right of=I1,yshift=0.5cm](WtK) {$\mathbf{w}_t^K$};
			\node[anchor=north,inner sep=.1cm,align=center,fill=white,right of=WtK] (How1) {MOOE};
			\node[MetaExpert, below of=How1, yshift=0.05cm](Wt) {$\mathbf{w}_t$};
			
			\node[OnlineExpert, below right of=I2, yshift=-0.5cm](WBK) {$\mathbf{w}_B^K$};
			\node[anchor=north,inner sep=.1cm,align=center,fill=white,right of=WBK] (How2) {$L_\mathcal{\widetilde{S}}^{\gamma}(\mathbf{w})$};
			\node[OfflineExpert, above of=How2](WK) {$\mathbf{w}^K$};
			
			\node[ExpertSet, right of=WK, xshift = 1cm] (ES){
				\node[OfflineExpert](W1) {$\mathbf{w}^1$};  \\ \node[OfflineExpert](W2) {$\mathbf{w}^2$}; \\ \node[outer sep = 3, rotate = 90](WS) {$\cdots$} ; \\ \node[OfflineExpert](WK1) {$\mathbf{w}^{K-1}$};\\};
			
			\draw[->,draw=black] (Ih1.east) -- (I1.west);
			\draw[->,draw=black] (Ih2.east) -- (Ih1.west);
			\draw[->,draw=black] (Ih3.east) -- (Ih2.west);
			\draw[->,draw=black,double,dashed] (I1.south) -- node [anchor=east] {Turn to} (I2.north);
			
			\draw[->,draw=black] (I1.north) -- node[fill=white]{Update} (WtK);
			\draw[->,draw=black] (WtK) -- (How1);
			\draw[->,draw=black] (ES) -- (How1.east);
			\draw[->,draw=black] (How1) -- node[fill=white]{Integrate} (Wt);
			\draw[->,draw=black] (Wt) -- node [near start, anchor=south] {Predict} (RS);
			
			\draw[->,draw=black] (I2.south) -- node[fill=white]{Update} (WBK);
			\draw[->,draw=black] (WBK) -- (How2);
			\draw[->,draw=black] (ES) -- (How2.east);
			\draw[->,draw=black] (How2) -- node[fill=white]{Minimize} (WK);
			\draw[->,draw=black] (WK) -- node [anchor=south,fill=white] {Refine} (ES);
			
			\matrix [draw=black,anchor=west,below of=Ih3, xshift = 0.6cm, yshift = -0.1cm,font=\footnotesize,minimum size=0.3cm, outer sep = 0, inner sep = 1.5]
			{
				\node[Sample,anchor=west, minimum size=0.3cm] {}; & \node[anchor=west] {: Sample}; \\
				\node[Sample,densely dotted,fill=red!20,rounded corners=.3ex,anchor=west, minimum size=0.3cm] {\qquad \quad}; & \node[anchor=west] {: Offline Interval}; \\
				\node[Sample,fill=green!20,rounded corners=.3ex,anchor=west, minimum size=0.3cm] {\qquad \quad}; & \node[anchor=west] {: Online Interval}; \\
				\node[OfflineExpert, anchor=west, minimum size=0.3cm] {} ; & \node[anchor=west] {: Offline Expert}; \\
				\node[OnlineExpert, anchor=west, minimum size=0.3cm] {} ; & \node[anchor=west] {: Online Expert}; \\
				\node[MetaExpert, anchor=west, dash pattern={on 2pt off 1pt},minimum size=0.3cm] {} ; & \node[anchor=west] {: Meta Expert}; \\
				\node[Sample,fill=cyan!20,loosely dotted,rounded corners=.3ex,anchor=west, minimum size=0.3cm] {\qquad \quad} ; & \node[anchor=west] {: Offline Expert Set}; \\
			};
			
			\begin{axis}
				[
				xticklabels={},
				yticklabels={},
				x=0.2cm,
				y = 1cm,
				line width=0.5mm,
				no markers,
				samples=50,smooth,
				axis x line*=bottom,
				axis y line*=left,
				enlargelimits=upper,
				yshift = -0.75cm,
				xshift = -13.7cm,
				above of=Ih3,
				anchor=east]
				\addplot {gauss(-2,0.5)};
			\end{axis}
			
			\begin{axis}
				[
				xticklabels={},
				yticklabels={},
				x=0.2cm,
				y = 1cm,
				line width=0.5mm,
				no markers,
				samples=50,smooth,
				axis x line*=bottom,
				axis y line*=left,
				enlargelimits=upper,
				yshift = -0.9cm,
				xshift = -6.1cm,
				above of=Ih1,
				anchor=east]
				\addplot {gauss(0,0.75)};
			\end{axis}
			
			\begin{axis}
				[
				xticklabels={},
				yticklabels={},
				x=0.2cm,
				y = 1cm,
				line width=0.5mm,
				no markers,
				samples=50,smooth,
				axis x line*=bottom,
				axis y line*=left,
				enlargelimits=upper,
				yshift = -0.95cm,
				xshift = -0.3cm,
				above of=I1,
				anchor=east]
				\addplot {gauss(2,1)};
			\end{axis}
		\end{tikzpicture}
	\end{adjustbox}
	\caption{The working process of MOOE on non-stationary time series.}
	\label{fig:mooe}
\end{figure*}

\section{Problem Statement}
In non-stationary time series, an online platform containing experts will receive an input \(\mathbf{x}\) at each time step and predict its label \(y\), indicating the class the input belongs to. For inputs with feedback, i.e., when the ground truth labels are revealed after predicting, the online platform will update by learning from the feedback. For inputs without feedback, the online platform merely predicts the labels. We aim to continuously update and utilize the experts in this online platform to more accurately predict class labels for samples from non-stationary time series in the current online interval.

Specifically, the considered non-stationary time series contains \(G - 1\) offline intervals and one online interval. Each offline interval contains \(B\) samples, and the online interval contains \(T\) samples (\(T \in [B]\)). We assume the distribution changes gradually, and the samples in each interval can be approximately drawn from a distribution. Accordingly, we set the maximal sample size \(B\) as a hyperparameter, even if the time between distribution changes is not constant and usually unknown. The online interval will become offline if \(T = B\), increasing the number of offline intervals. Accordingly, we have the following assumptions.
\begin{assumption}
	\label{as:1}
	Let $\mathcal{D}_g$ be the data distribution in the $g^{\text{th}}$ interval. $\mathcal{D}_{\mathcal{U}} = \bigcup_{g = 1}^G \mathcal{D}_g$ is non-stationary since $\mathcal{D}_g \neq \mathcal{D}_{g'}, \forall g,g' \in [G], g \neq g'.$
\end{assumption}

\begin{assumption}
	\label{as:3} The norm of every input sample $\mathbf{x}$ with label $y$ in the Hilbert space  i.i.d. drawn from the distribution $\mathcal{D}_{G}$ of the online interval is upper bounded by a constant $D$:
	\begin{equation*}
		\Vert \mathbf{x}\Vert \leq D, \forall (\mathbf{x},y) \thicksim \mathcal{D}_{G}.
	\end{equation*}
	The eigendecomposition of the Hilbert-Schmidt operator is
	\begin{equation*}
		\mathbb{E}_{(\mathbf{x},y) \thicksim \mathcal{D}_{G}}[\mathbf{x} \mathbf{x}^T]  = \sum_{i = 1}^\infty \lambda_i \mathbf{u}_i\mathbf{u}_i^T,
	\end{equation*}
	where $(\mathbf{u}_i)_{i = 1}^\infty$ forms an orthonormal basis of Hilbert Space and $(\lambda_i)_{i = 1}^\infty$ corresponds to the eigenvalues in a non-increasing order.
\end{assumption}
\begin{assumption}
	\label{as:2}
	For any sample $(\mathbf{x},y) \thicksim \mathcal{D}_{\mathcal{U}}$, the hypothesis class is
	\begin{equation*}
		\mathcal{H} \triangleq \{h: \mathbf{x} \mapsto \left< \mathbf{w}, \mathbf{x}\right> \, | \, \mathbf{w} \in \mathcal{W} , \Vert \mathbf{w}\Vert \leq R\},
	\end{equation*}
	where the domain $\mathcal{W}$ bounded by $R$ is a convex subspace of a Hilbert space.
\end{assumption}
\begin{assumption}
	\label{as:4}
	For any sample $(\mathbf{x},y) \thicksim \mathcal{D}_{\mathcal{U}}$, the loss function family $\mathcal{L}$ with the hypothesis class $\mathcal{H}$ is bounded in the interval $[0,1]$:
	\begin{equation*}
		\mathcal{L} \triangleq \{ (\mathbf{x},y) \mapsto l(h(\mathbf{x}),y ) \, | \, h \in \mathcal{H}, l(h(\mathbf{x}),y ) \in [0,1] \}.
	\end{equation*}
\end{assumption}
\begin{assumption}
	\label{as:5}
	For any $(\mathbf{x},y) \thicksim \mathcal{D}_{\mathcal{U}}$ and all $\mathbf{w},\mathbf{w}' \in \mathcal{W}$, $l(\left<\cdot,\mathbf{x} \right>,y ) $ is convex and $\beta$-smooth over the domain $\mathcal{W}$:
	\begin{equation*}
		\left\Vert \nabla l(\left<\mathbf{w},\mathbf{x} \right>,y ) - \nabla l(\left<\mathbf{w}',\mathbf{x} \right>,y )\right\Vert \leq \beta \left\Vert \mathbf{w} - \mathbf{w}'  \right\Vert.\\
	\end{equation*}
\end{assumption}

In the $G^{\text{th}}$ interval, we would like to learn an expert $\mathbf{w} \in \mathcal{W}$ with a small popular risk with respect to the nonnegative loss function $l$
\begin{equation}
	\label{PR}
	L_\mathcal{D}(\mathbf{w}) = \mathbb{E}_{(\mathbf{x},y) \thicksim \mathcal{D} } [l(\left<\mathbf{w},\mathbf{x} \right>,y )],
\end{equation}
by minimizing the corresponding empirical risk using the proposed method:
\begin{equation}
	\label{ER}
	L_\mathcal{S}(\mathbf{w}) = \frac{1}{T} \sum_{t = 1}^T l(\left<\mathbf{w},\mathbf{x}_t \right>,y_t ) = \frac{1}{T} \sum_{t = 1}^T f_t(\mathbf{w}),
\end{equation}
where $\mathcal{S} = \{ (\mathbf{x}_1, y_1),\ldots,(\mathbf{x}_T, y_T)\}$ is the data set consisting of $T (T \in [B])$ samples in the online interval, and we use $L_{\mathcal{\widetilde{S}}}(\mathbf{w})$ to denote the specific case when $T = B$. Let $\mathbf{w}^* \in \arg \min_{w \in \mathcal{W}} L_\mathcal{D}(\mathbf{w})$ be an optimal solution and $\widehat{\mathbf{w}} \in \arg \min_{w \in \mathcal{W}} L_\mathcal{S}(\mathbf{w})$ be an empirical minimizer.


Because the loss function $l$ is nonnegative as well as $\beta$-smooth, according to the self-bounding property~\citep{SM:10} of smooth functions and~\cref{as:4}, we obtain the following upper bound on the norm of the gradients of $l(\left<\cdot,\mathbf{x} \right>,y )$ for any $(\mathbf{x},y) \thicksim \mathcal{D}_{\mathcal{U}}$ and all $\mathbf{w} \in \mathcal{W}$:
\begin{equation}
	\begin{aligned}
		\label{eq:func}
		\Vert \nabla l(\left<\mathbf{w},\mathbf{x} \right>,y ) \Vert \leq \sqrt{4 \beta \cdot l(\left<\mathbf{w},\mathbf{x} \right>,y )} \leq 2\sqrt{\beta}.
	\end{aligned}
\end{equation}
Note that this paper aims to address the non-stationary issue rather than the widely-explored non-convex problem. We thus assume the loss function is convex for convenience and focus on providing the theoretical guarantees for the proposed learning mechanism.

\section{Mixture of Online and Offline Experts}
\cref{fig:mooe} introduces the working process of Mixture of Online and Offline Experts (MOOE) for the non-stationary time series. MOOE maintains several offline experts for the corresponding offline intervals and an online expert for the current online interval. It then integrates all of these experts using a meta expert with adaptive weights.

The number of maintained experts is $K$, which is defined as,
\begin{equation}
	K = \begin{cases}
		K_{\text{max}},  & \text{if $G \geq K_{\text{max}}$} \\
		G, & \text{if $G < K_{\text{max}}$}
	\end{cases}.
\end{equation}
where $K_{\text{max}}$ is a hyperparameter denoting the maximal number of maintained experts. Therefore, MOOE contains $K-1$ offline experts and one online expert. In the interest of brevity, an expert and its corresponding advice are denoted as its parameters $\mathbf{w}$. Accordingly,  we assume the $k^{\text{th}}~(k \in [K - 1])$ offline expert is $\mathbf{w}_t^k$ and the online expert is $\mathbf{w}_t^K $. For the $t^{\text{th}}$ sample with feedback in the online interval, MOOE firstly selects $K$ experts
\begin{equation}
	\{\underbrace{\mathbf{w}^1_t, \ldots, \mathbf{w}^{K - 1}_t}_{\text{Offline Expert}}, \underbrace{\mathbf{w}^K_t}_{\text{Online Expert}} \},
\end{equation}
and integrates them into a meta expert $\mathbf{w}_t$ for making a predition. When $T = B$, the online interval becomes offline, and a new online interval appears. We generate the new offline expert $\mathbf{w}^K$ for the just-passed complete online interval and refresh $K - 1$ offline experts if $G \geq K_{\text{max}}$.

\subsection{Meta Expert}
The meta expert adjusts its strategy of integrating the $K$ experts ($K - 1$ offline experts and one online expert) according to their losses received on labeled samples. For the online interval, we track the best expert~\cite{TB:95} based on the exponentially weighted average forecaster~\cite{PLG:06} by assigning a considerable weight to the expert with a small cumulative loss, and vice verse. Accordingly,  at iteration $t$ in the online interval, the meta expert outputs a weighted average solution
\begin{equation}
	\label{eq:wf}
	\mathbf{w}_t = \sum_{k = 1}^{K - 1} \alpha_t^k \mathbf{w}^k_t + \alpha_t^K \mathbf{w}^K_t = \sum_{k = 1}^{K} \alpha_t^k \mathbf{w}^k_t,
\end{equation}
$\alpha_t^k$ is the weight of the $k^{\text{th}}$ expert $\mathbf{w}^k_t$. To lead to a compact regret bound, ensure that $\sum_{k = 1}^K \alpha_1^k = 1$, and provide different weights for experts according to their priorities, $\alpha_t^k$ is initialized as
\begin{equation}
	\label{eq:in}
	\alpha_1^k = \frac{K + 1}{(K + 1 -k)(K + 2 -k)K}.
\end{equation}
Note that it is unnecessary to project $\mathbf{w}_t$ into the domain $\mathcal{W}$. Because each expert satisfies $\mathbf{w}^k_t \in \mathcal{W} (k \in [K])$ and the weighting function Eq.~(\ref{eq:wf}) is linear, the weighted average $\mathbf{w}_t$ is still in the domain $\mathcal{W}$ according to convex properties.

After obtaining the loss at iteration $t$, the $K$ weights are updated according to the exponential weighting scheme
\begin{equation}
	\label{eq:update:a}
	\alpha^k_t = \frac{\alpha^k_t e^{-\nu f_t(\mathbf{w}^{k}_t)}}{\sum_{k' = 1}^{K} \alpha^{k'}_t e^{-\nu f_t(\mathbf{w}^{k'}_t)}},
\end{equation}
where $\nu = 4\sqrt{\frac{\ln K}{T}}$ is the step size.  MOOE is summarized in~\cref{alg:algorithm}.

\begin{algorithm}
	\caption{MOOE}
	\label{alg:algorithm}
	\begin{algorithmic}[1]
		\STATE {\textbf{Input:} step size $\nu$ \\
			\quad \quad \quad online expert $\mathbf{w}_1^{K}$ \\
			\quad \quad \quad offline expert set $\{\mathbf{w}^1, \ldots, \mathbf{w}^{K - 1}\}$}
		\STATE Initialize $\alpha^1_1 < \alpha^2_1 < \cdots < \alpha^K_1$ according to:
		\begin{equation*}
			\alpha_1^k = \frac{K + 1}{(K + 1 -k)(K + 2 -k)K}, \forall k \in [K]
		\end{equation*}
		\FOR{t = 1,\ldots,T}
		\STATE Receive online expert $\mathbf{w}^K_t$
		\STATE Assign offline expert $\mathbf{w}^k_t = \mathbf{w}^k, \forall k \in [K - 1]$
		\STATE Output weighted average: $\mathbf{w}_t = \sum_{k = 1}^{K} \alpha_t^k \mathbf{w}^k_t$
		\STATE Receive the loss function $f_t(\cdot)$
		\STATE Update expert weights:\\
		\begin{equation*}
			\alpha^k_t = \frac{\alpha^k_t e^{-\nu f_t(\mathbf{w}^{k}_t)}}{\sum_{k' = 1}^{K} \alpha^{k'}_t e^{-\nu f_t(\mathbf{w}^{k'}_t)}}, \forall k \in [K]
		\end{equation*}
		\STATE Send gradient $\nabla f_t(\mathbf{w}^{K}_t)$ to the online expert
		\ENDFOR
	\end{algorithmic}
\end{algorithm}

\subsection{Offline Expert}
We extract knowledge from offline intervals by learning an offline expert for each online interval when all of its samples are available. Each interval is coupled with its previous offline experts and online expert when its online expert has passed this interval once, and its previous offline experts may be learned from similar distributions. Thus, we transfer their knowledge adaptively to its offline expert.

For each online interval, we calculate a new offline expert $\mathbf{w}^{K}$ once $T = B$ by taking advantage of the prior knowledge of the $K-1$ offline experts $\mathbf{w}^1_B, \ldots, \mathbf{w}^{K - 1}_B$ and the online expert $\mathbf{w}^K_B$. According to the strategy of the meta expert, the expert performing best in this interval has the largest weight, therefore the new offline expert $\mathbf{w}^{K}$ should be close to the best expert. Accordingly, we use the regularization term $\Omega(\mathbf{w}) = \left\Vert\mathbf{w} - \sum_{k = 1}^K \alpha_B^k \mathbf{w}^k_B \right\Vert_2^2$ to constrain the search space of $\mathbf{w}^{K}$,
\begin{equation}
	\label{eq:off}
	\mathbf{w}^K = \arg \min_{w \in \mathcal{W}} \underbrace{\frac{1}{B} \sum_{t = 1}^B f_t(\mathbf{w}) + \frac{\gamma}{2} \Omega(\mathbf{w})}_{L_\mathcal{\widetilde{S}}^{\gamma}(\mathbf{w})},
\end{equation}
where $\gamma \geq \sum_{k = 1}^K \alpha_B^k L_\mathcal{\widetilde{S}}(\mathbf{w}_B^k) /4 R^2$ is a hyperparameter to control the effect of the prior knowledge and $T$ is assigned as $B$ in $\mathcal{\widetilde{S}}$.

After receiving the new offline expert $\mathbf{w}^K$, we set priorities for all $K$ offline experts, as their potential abilities for the next online interval vary. We then select $K-1$ offline experts by eliminating the one with the lowest priority and initialize their weights in the meta expert according to these priorities, as shown in Eq.~(\ref{eq:in}). The priority mechanism does not affect our theoretical results for the MOOE method, so we do not delve into it here. Instead, we provide two simple mechanisms: maintaining an expert queue where the first expert has the lowest priority and the newest expert is enqueued while the oldest is removed, or setting the newest offline expert with the highest priority and assigning the priorities for the previous $K-1$ experts based on their weights.

\subsection{Online Expert}
To train an online expert for a new online interval, we can reinitialize its parameters randomly or by inheriting its solution from the just-passed complete online interval as a warm start. Recall that we can train the online expert by any off-the-shelf online optimization methods on the fly. In this paper, we use the standard Online Gradient Descent (OGD)~\cite{OSG:03} method as an instance because it is the most common and famous online optimization method. On the online interval, the online expert submits its advice $\mathbf{w}^{K}_t$ to the meta expert and receives the gradient $\nabla f_t(\mathbf{w}^{K}_t)$ to update its parameters by
\begin{equation}
	\mathbf{w}^{K}_{t + 1} = \Pi_{\mathcal{W}} [ \mathbf{w}^{K}_t - \eta_t \nabla f_t(\mathbf{w}^{K}_t) ]
\end{equation}
$\eta_t = \frac{D}{\sqrt{\beta t}}$ is the step size, and $\Pi_{\mathcal{W}}$ is the proximal operator onto space $\mathcal{W}$.

\section{Theoretical Guarantees}
In this section, we provide theoretical guarantees for MOOE, which match our expectations. Specifically, we analyze the properties of the regularization term $\Omega(\mathbf{w})$ and provide the regret and the generalization error of the output hypothesis. To exploit the convexity, smoothness, and nonnegativity conditions of the loss function, the hypothesis class, the data distribution, and the regret, we involve the data-independent excess risk of $\widehat{\mathbf{w}}$, the Rademacher complexity of hypothesis class $\mathcal{H}$ w.r.t. $\mathcal{D}$ and the regret for implying the generalization. 

\subsection{Parameter Convergence}
The hyperparameter $\gamma$ for $\Omega(\mathbf{w})$ should be assigned with considerable value to ensure the validity of the regularization. To process, we derive the upper bound of this regularization.
\begin{lemma}
	\label{le:hy1}
	$L_\mathcal{\widetilde{S}}^{\gamma}(\mathbf{w})$ is strongly-convex w.r.t. $\mathbf{w} \in \mathcal{W}$, and
	\begin{equation*}
		\Omega(\mathbf{w}^K)  \leq  \sum_{k = 1}^K \alpha_T^k L_{\mathcal{\widetilde{S}}}(\mathbf{w}_B^k) / \gamma.
	\end{equation*}
	We set $\gamma \geq \left( \sum_{k = 1}^K \alpha_B^k L_{\mathcal{\widetilde{S}}}(\mathbf{w}_B^k) \right) / 4 R^2$ to ensure the validity of the regularization term.
\end{lemma}
Accordingly, the following theorem shows the benefit of this regularization, it can narrow the gap between the minimizer $\mathbf{w}^K$ and the optimal solution $\mathbf{w}^*$ by applying the maintained $K$ experts adaptively and setting $\gamma$ carefully.
\begin{theorem}
	\label{th:hy2}
	By setting $\gamma \geq \left( \sum_{k = 1}^K \alpha_B^k L_{\mathcal{\widetilde{S}}}(\mathbf{w}_B^k) \right) / 4 R^2$  and using $\mathbf{w}_B^1,\mathbf{w}_B^2,\ldots,\mathbf{w}_B^K$ as prior knowledge to obtain $\mathbf{w}^K$ from $L_{\mathcal{\widetilde{S}}}^{\gamma}(\mathbf{w})$, we have
	\begin{align*}
		\left\Vert \mathbf{w}^K - \mathbf{w}^*\right\Vert \leq \sqrt{2 \Omega(\mathbf{w}^*)  + \frac{32\beta}{\gamma^2} + \frac{6 \sum_{k = 1}^K \alpha_B^k L_{\mathcal{\widetilde{S}}}(\mathbf{w}_B^k)}{\gamma}}.
	\end{align*}
\end{theorem}
Although it is impossible for us to obtain $\mathbf{w}^*$ since the distribution for this interval is unknown, we can obtain an approximate solution by using the regularization term $\Omega(\mathbf{w})$. If the optimal solution is close to the weight average $\sum_{k = 1}^K \alpha_T^k \mathbf{w}^k_T$, the value of $\Omega(\mathbf{w}^*)$ and the  upper bound of the difference $\left\Vert \mathbf{w}^K - \mathbf{w}^*\right\Vert$ are small. Although it is also impossible for us to measure $\Omega(\mathbf{w}^*)$, we can measure the weighted term $\sum_{k = 1}^K \alpha_B^k L_{\mathcal{\widetilde{S}}}(\mathbf{w}_B^k)$ in the above upper bound where $L_{\mathcal{\widetilde{S}}}(\mathbf{w}_B^k)$ is the empirical error of the $k^{\text{th}}$ expert in the latest interval. As a result, we know that the empirical minimizer $\mathbf{w}^K$ of $L_{\mathcal{\widetilde{S}}}^{\gamma}(\mathbf{w})$ approaches the optimal solution $\mathbf{w}^*$ of the original problem $L_\mathcal{D}(\mathbf{w})$ if these experts considered in the regularization term $\Omega(\mathbf{w})$ are effective in the latest interval. To sharpen this bound, the weights for experts with small empirical errors should be larger and the design of the meta expert can meet the need. Therefore, we can draw a conclusion that $\mathbf{w}^K$ should be close to these experts with small empirical errors in the domain $\mathcal{W}$. This conclusion leads to the design of the regularization term $\Omega(\mathbf{w})$.

\subsection{Regret Bound}
The following regret measures the performance of MOOE
\begin{equation}
	\text{Regret}_{\text{MOOE}} = \sum_{t = 1}^T f_t(\mathbf{w}_t) - \min_{\mathbf{w} \in \mathcal{W}} \sum_{t = 1}^T f_t(\mathbf{w}).
\end{equation}
However, it is hard to minimize the regret directly because the output $\mathbf{w}_t$ is related to a meta expert, an online expert, and $K - 1$ offline experts. Therefore, we decompose the regret into two regrets:  $\text{Regret}_\text{ME}$ w.r.t. the meta expert and  $\text{Regret}_\text{KE}$ w.r.t. online and offline experts. Further, we can bound $\text{Regret}_\text{KE}$ by $\text{Regret}_\text{OE}$ which corresponds to the online expert. Therefore, we can obtain the regret bound of $\text{Regret}_{\text{MOOE}}$ by bounding $\text{Regret}_\text{ME}$ and $\text{Regret}_\text{OE}$ separately.
\begin{equation}\label{th:sp}
	\begin{aligned}
		\text{Regret}_{\text{MOOE}} & = \text{Regret}_\text{ME} + \text{Regret}_\text{KE} \\
		\leq & \text{Regret}_\text{ME} + \text{Regret}_\text{OE},
	\end{aligned}
\end{equation}
where
\begin{equation}
	\begin{aligned}
		\text{Regret}_\text{ME} & = \sum_{t = 1}^T f_t(\mathbf{w}_t) -  \min_{k \in [K]} \sum_{t = 1}^T f_t(\mathbf{w}^k_t), \\\text{Regret}_\text{KE} & = \min_{k \in [K]} \sum_{t = 1}^T f_t(\mathbf{w}^k_t) - \sum_{t = 1}^T f_t(\widehat{\mathbf{w}}), \\\text{Regret}_\text{OE}  & = \sum_{t = 1}^T f_t(\mathbf{w}^K_t) - \sum_{t = 1}^T f_t(\widehat{\mathbf{w}}).
	\end{aligned}
\end{equation}
The online expert $\mathbf{w}^K_t$ never surpasses that of the best expert among all the $K$ experts because it is also one of them. Besides, it is impossible to obtain the regret for the offline experts since they are pre-given and their parameters do not change after receiving the loss $f_t(\cdot)$. Specifically, we have the following theorem.
\begin{theorem}
	\label{th:rg}
	The MOOE method with step sizes $\{ \nu = 4\sqrt{\frac{\ln K}{T}}, \eta_t = \frac{D}{\sqrt{\beta t}}, t \in [T] \}$ guarantees the following regret for all $ 1 \leq T \leq B$,
	\begin{align*}
		\text{Regret}_{\text{MOOE}} \leq \sqrt{T \ln K} + 6D\sqrt{T\beta},
	\end{align*}
	and the number of experts $K$ and samples $T$ should satisfy
	\begin{align*}
		K \leq 2 \exp \left(6D \sqrt{\beta} - \frac{\text{Regret}_\text{KE}}{\sqrt{T}} \right),
	\end{align*}
	to ensure that the advice from MOOE gives an equivalent or better result than that from its online expert.
\end{theorem}
Accordingly, the regret of MOOE for the online interval is $O(\sqrt{T})$, which is consistent with that of the chosen online expert. However, MOOE works better, i.e. $\text{Regret}_{\text{MOOE}} \leq \text{Regret}_{\text{OE}}$, if $K$ and $T$ satisfy the condition in Theorem~\ref{th:rg}. In theory, we have $\text{Regret}_\text{KE} \leq \text{Regret}_\text{OE} \leq 6D\sqrt{T\beta}$. These offline experts are better than the online expert when their corresponding data distributions are approximately matched, or the number of observed labeled samples in the current interval is limited. The first inequality is strict, and $K$ is bounded by a positive value. On the other hand, the number of samples in an interval $T \leq B$ should not be too large. Although the bound of $K$ depends on $\text{Regret}_\text{KE}$, it is impossible to bound this term without any further assumptions because the $K$ experts are trained from different data sets. Fortunately, it is unnecessary to set K strictly according to its conditions. We can apply MOOE if we believe that the regret of the best offline expert can surpass that of the online expert at least $6D\sqrt{T\beta} - \text{Regret}_\text{KE} = \sqrt{T \ln K}$. The assumption is mild since we can set a small K (like 2 or 3) even without prior knowledge. An intuitive understanding is that: if $K$ is too large, it is difficult for the meta expert to derive effective advice because of the dilution effect from those weak experts; if $B$ is too large, the samples in an interval may come from various distributions, and the assumption about the setting may not hold.

\subsection{Generalization Error Bound}
The MOOE performance is measured by the excess risk $L_\mathcal{D}(\overline{\mathbf{w}}) - L_\mathcal{D}(\mathbf{w}^* )$ where $\overline{\mathbf{w}} = \frac{1}{T} \sum_{t = 1}^T \mathbf{w}_t$ is the average of the online interval. To derive an algorithmic bound, we introduce the intermediate term $L_\mathcal{D}(\widehat{\mathbf{w}})$ because $\widehat{\mathbf{w}}$ as an empirical minimizer of $L_\mathcal{S}(\widehat{\mathbf{w}})$ is necessary for analyzing the regret. Taking the divide-and-conquer approach, we have
\begin{equation}
	\begin{aligned}
		\label{eq:all:1}
		& L_\mathcal{D}(\overline{\mathbf{w}}) - L_\mathcal{D}(\mathbf{w}^*) \leq  \frac{1}{T} \sum_{t = 1}^T L_\mathcal{D}(\mathbf{w}_t) - L_\mathcal{D}(\mathbf{w}^*) =\\
		& \underbrace{\frac{1}{T} \sum_{t = 1}^T L_\mathcal{D}(\mathbf{w}_t) - L_\mathcal{D}(\widehat{\mathbf{w}})}_{\mathcal{B}_1} + \underbrace{L_\mathcal{D}(\widehat{\mathbf{w}}) - L_\mathcal{D}(\mathbf{w}^*)}_{\mathcal{B}_2}.
	\end{aligned}
\end{equation}
The inequality is owing to the convexity of $L_\mathcal{D}(\cdot)$, which implies $ L_\mathcal{D}( \frac{1}{T} \sum_{t = 1}^T \mathbf{w}_t) \leq \frac{1}{T} \sum_{t = 1}^T L_\mathcal{D}(\mathbf{w}_t)$. The regret of MOOE is applied to imply the upper bound of $\mathcal{B}_1$ by the following lemma.
\begin{lemma}
	\label{le:erm:rg}
	Following \cref{th:rg}, with probability at least $1 - \delta$, we have
	\begin{equation*}
		\begin{aligned}
			\frac{1}{T} \sum_{t = 1}^T L_\mathcal{D}(\mathbf{w}_t) - L_\mathcal{D}(\widehat{\mathbf{w}}) \leq \frac{\sqrt{\ln K} + 6D\sqrt{\beta} + 4 \log(4 / \delta)}{\sqrt{T}}.\\
		\end{aligned}
	\end{equation*}
\end{lemma}
Following the advanced study for any norm-regularized hypothesis class~\cite{LRC:18} and the self-bound property of smooth functions~\cite{SM:10}, we can derive the following data-dependent generalization bound for $\mathcal{B}_2$ by the following lemmas.

\begin{lemma}
	\label{le:erm:di}
	Exploiting the convexity, smoothness, and nonnegativity conditions of the loss function family $\mathcal{L}$, with probability at least $1 - \delta$, $L_\mathcal{D}(\widehat{\mathbf{w}}) - L_\mathcal{D}(\mathbf{w}^*)$ is bounded by
	\begin{equation*}
		\frac{\left(12\beta R^2 + 4R \sqrt{\beta} \right)\log(4 / \delta) }{T} +  4R \sqrt{\frac{2 \beta \log(4/\delta)}{T}}.
	\end{equation*}
\end{lemma}

\begin{lemma}
	\label{le:erm:dd}
	Exploiting the hypothesis class $\mathcal{H}$ and the distribution $\mathcal{D}$ of the observed data at the online interval, with probability at least $1 - \delta$, $L_\mathcal{D}(\widehat{\mathbf{w}}) - L_\mathcal{D}(\mathbf{w}^*)$ is bounded by
	\begin{equation*}
		\begin{aligned}
			42\sqrt{6 \beta} \log^{\frac{3}{2}}(64T) \mathcal{R}_{\mathcal{D}}(\mathcal{H}) + 3 \sqrt{\frac{\log(4 / \delta)}{T}}.
		\end{aligned}
	\end{equation*}
\end{lemma}
where $ \mathcal{R}_{\mathcal{D}}(\mathcal{H})$ is the Rademacher complexity of hypothesis space $\mathcal{H}$.

Using the excess risk bound framework in Eq.~(\ref{eq:all:1}), we obtain the following generalization error bound by considering \cref{le:erm:rg},~\cref{le:erm:di} and~\cref{le:erm:dd}.
\begin{theorem}
	\label{th:all}
	Exploiting the loss function properties (convexity, smoothness, and nonnegativity) of $\mathcal{L}$, the hypothesis class $\mathcal{H}$, the data distribution $\mathcal{D}$ and the regret of MOOE, with probability at least $1 - \delta$, we have
	\begin{equation*}
		\begin{aligned}
			& L_\mathcal{D}(\overline{\mathbf{w}}) - L_\mathcal{D}(\mathbf{w}^*) \leq  \frac{\left(12\beta R^2 + 4R \sqrt{\beta} \right)\log(16 / \delta) }{T}  \\
			&  + \frac{28R\sqrt{\beta} \log^{\frac{3}{2}}(64T)}{T} \left(\sqrt{ \sum_{i = 1}^{\infty} \left( T D^2 \wedge e \lambda_i \right)} +  D\sqrt{e}\right) \\
			& + \frac{\left( 6(R + D) \sqrt{\beta} + 2\right) \sqrt{\log(16/\delta)} + 4 \log(8 / \delta) + \sqrt{\ln K}}{\sqrt{T}}. \\
		\end{aligned}
	\end{equation*}
\end{theorem}
The convergence rate for the generalization error is $O(1 / \sqrt{T})$, which is consistent with that in stationary and non-algorithmic cases~\cite{CO:08}. The bound reflects the best result achieved so far without any other assumptions. Furthermore, it is directly related to the sample complexity, and the result is algorithmic. This result triggers an immediate problem: Can we use fewer samples to achieve a desirable generalization error if the used off-the-shelf online optimization method can achieve a better regret? Unfortunately, the answer is not affirmative. The intuition behind the problem is that the bottleneck is not on the optimization method.

\begin{figure*}[th]
	\includegraphics[width=0.16\textwidth]{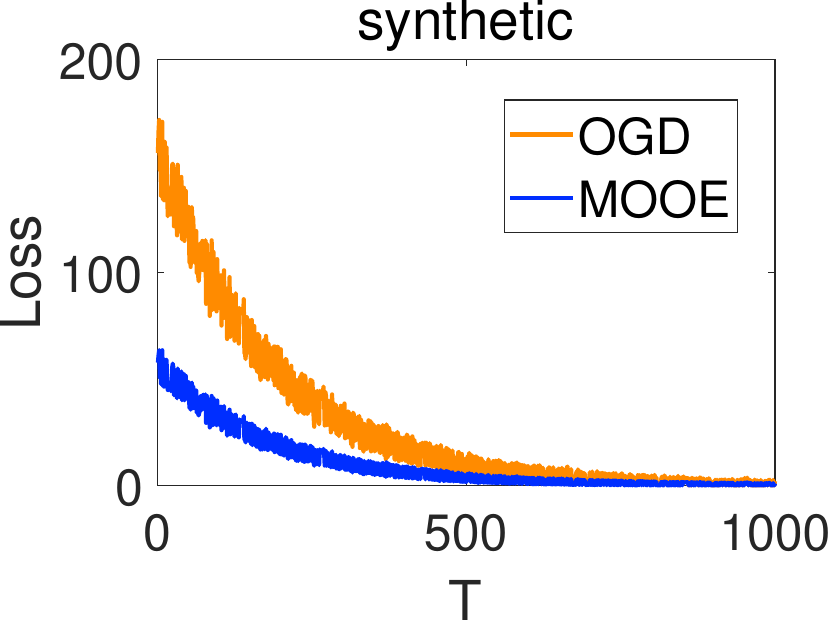}
	\includegraphics[width=0.16\textwidth]{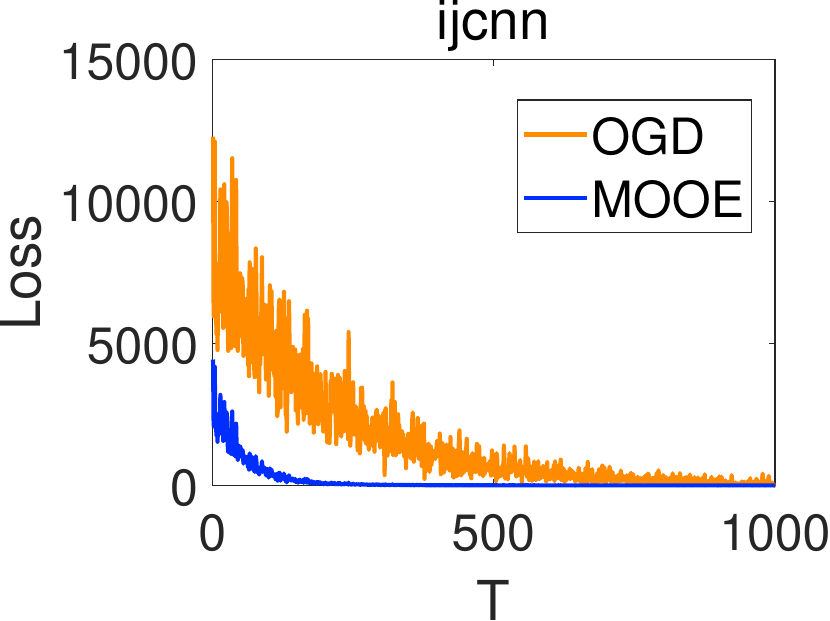}
	\includegraphics[width=0.16\textwidth]{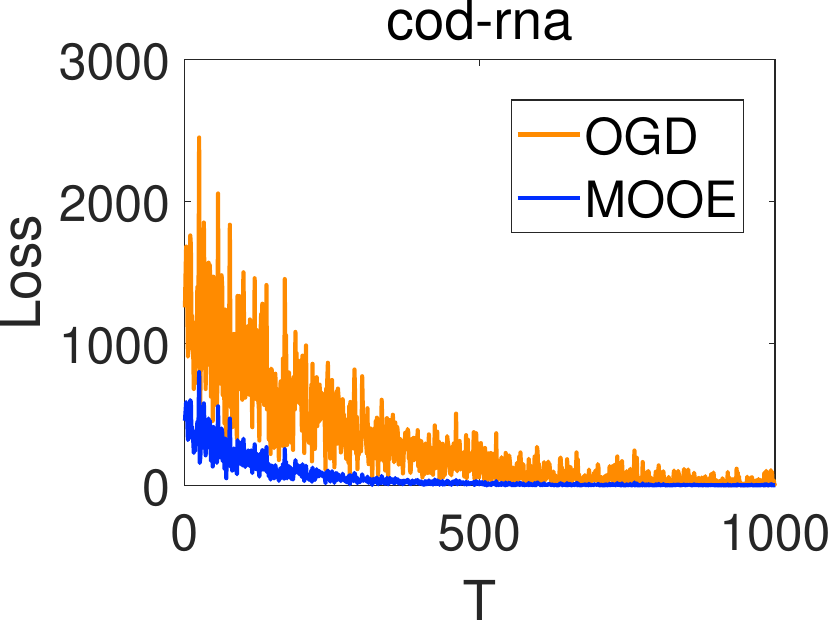}
	\includegraphics[width=0.16\textwidth]{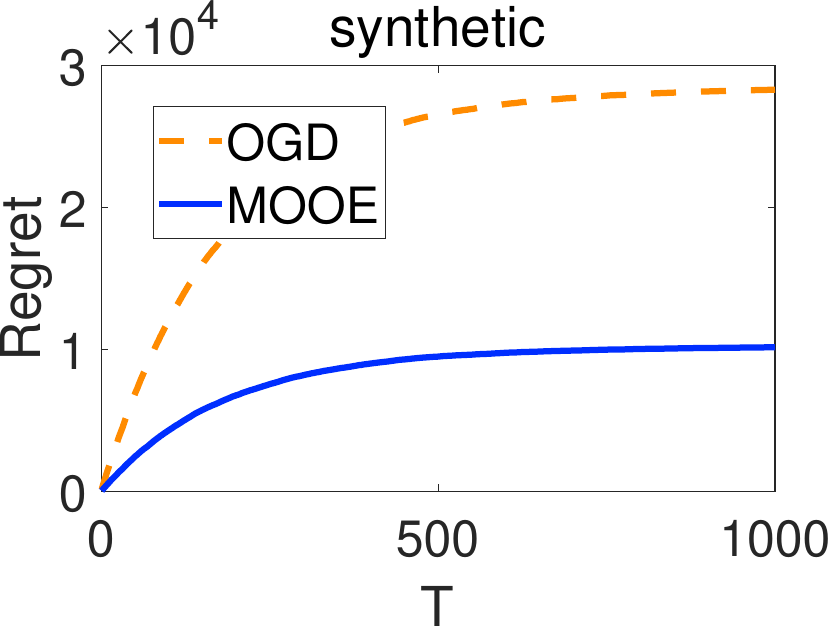}
	\includegraphics[width=0.16\textwidth]{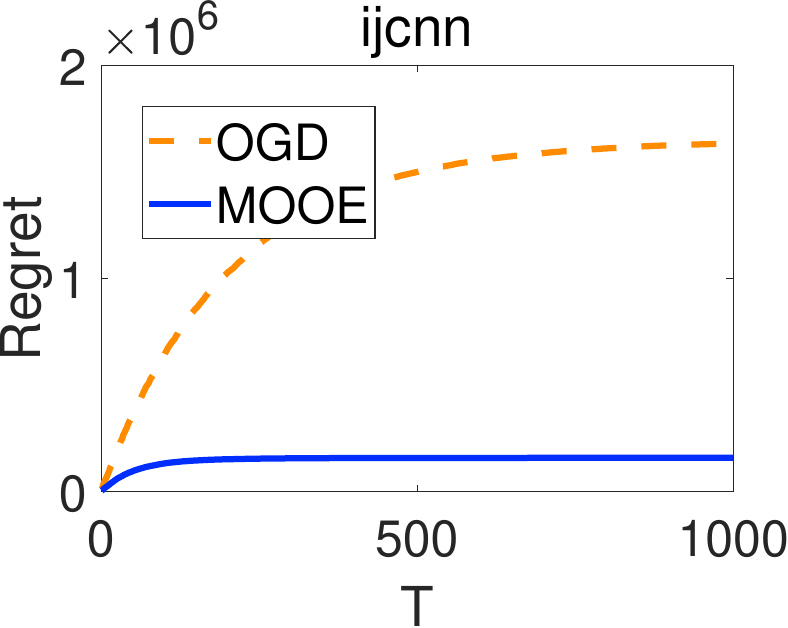}
	\includegraphics[width=0.16\textwidth]{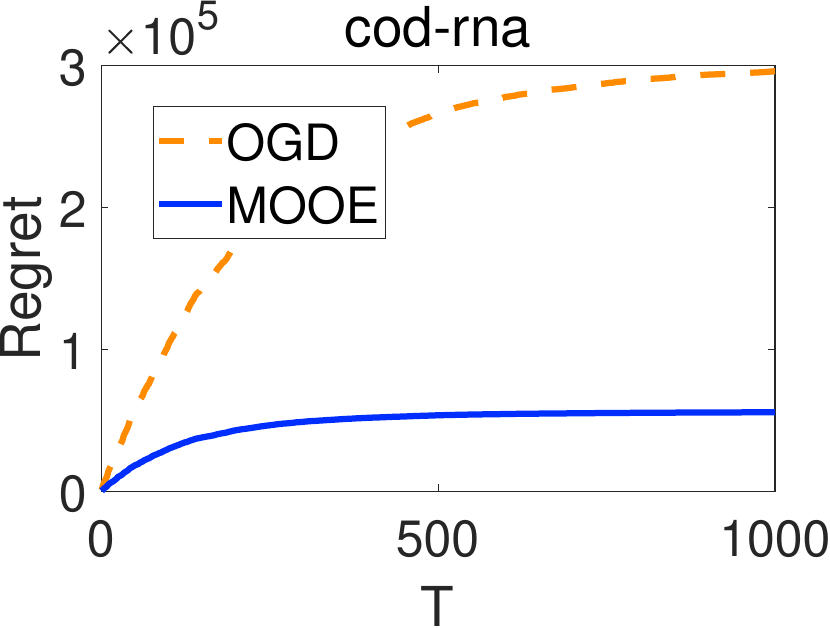}
	\caption{Regret and loss of MOOE and OGD methods.}
	\label{fig:res}
\end{figure*}

\begin{table}[t] 
	\centering
	\begin{tabular}{cccccc}
		\toprule
		Dataset & NSE & DTEL & Condor & MOOE \\ \midrule  \midrule
		Usenet & 63.8 & 68.0 & 73.1 & \bf{78.5} \\
		Weather & 76.0 & 68.9 & 79.4 & \bf{82.4} \\
		GasSensor & 42.4 & 63.8 & 81.6 & \bf{83.1} \\
		Powersupply & 74.0 & 69.9 & 72.8 & \bf{77.9} \\
		Electricity & 79.0 & 81.0 & 84.7 & \bf{85.6} \\
		Covertype & 79.0 & 69.4 & 89.6 & \bf{90.0} \\
		WESAD & 70.4 & 73.8 & 86.3 & \bf{89.9} \\
		Kitsune & 73.9 & 71.6 & 87.3 & \bf{93.4} \\ \bottomrule
	\end{tabular}
	\caption{Comparisons on real-world data.}
	\label{tb:comp}
\end{table}

\section{Experimental Results}~\label{sec:ea}
In this section, we present empirical analysis to support our proposed theory and model\footnote{The source code is publicly available at: \url{https://github.com/Lawliet-zzl/MOOE}.}. 


\subsection{Regret on Synthetic and Real-world Datasets}
We address binary classification on non-stationary time series and compare MOOE with OGD using both synthetic and real-world datasets (ijcnn and cod-rna) from the LIBSVM repository~\cite{LIBSVM:11}. On the synthetic dataset, each interval features samples from two-dimensional Gaussian distributions with dynamically changing means. For the synthetic datasets, we divide the data into 15 intervals, applying Gaussian noise to simulate dynamic changes. We maintain a maximum of five experts in MOOE to ensure fairness in comparison. Theoretical analyses show that MOOE outperforms OGD in dynamic environments, maintaining a convergence rate of $O(1 / \sqrt{T})$. As shown in~\cref{fig:res}, Empirical results indicate that MOOE achieves significantly lower loss than OGD, particularly at the early stages with few samples, due to the integration of offline expert knowledge. Additionally, MOOE exhibits smaller regret over time, adapting effectively to samples by adjusting the strategy of integrating offline and online experts.
\subsection{Predictive Accuracy}
\subsubsection{Real-world Non-stationary Time Series}
To verify the effect of the proposed MOOE method, we perform comparison experiments following the setup~\cite{MR:18}. Specifically, we use eight real-world non-stationary time series datasets, including Usenet~\cite{U:08}, Weather~\cite{NSE:11}, GasSensor~\cite{GAS:12}, Powersupply~\cite{UCR:19}, Electricity~\cite{ELE:99} ,Covertype~\cite{DTEL:18}, WESAD~\cite{WESAD:18}, and Kitsune~\cite{Kitsune:18}. We compare MOOE with three state-of-the-art methods, including NSE~\cite{NSE:11}, DTEL~\cite{DTEL:18}, and Condor~\cite{MR:18}. In the experiments, we adopt the maximum sample size of an interval  $B = 50$ and the maximal number of maintained experts $K_{\text{max}} = 25$. The overall mean of predictive accuracy is reported, which indicates the average performance of the algorithm over the whole time series. The comparison results are reported in~\cref{tb:comp}. The results show that the proposed MOOE method outperforms other contenders. Specifically, MOOE achieves $ 13.7 \%$ improvement over the other state-of-the-art methods. This is because MOOE can utilize the knowledge of the offline experts to adopt each new online interval. These experimental results show the superiority of the proposed MOOE method.

\begin{table} \scriptsize
	\centering
	\setlength{\tabcolsep}{5pt}
	\begin{tabular}{lcccccc}
		\toprule
		\multirow{2}{*}{Algorithm} & \multicolumn{3}{c}{Email list} & \multicolumn{3}{c}{Spam filtering} \\ \cmidrule(lr){2-4} \cmidrule(lr){5-7}
		& Accuracy & Precision & Recall   & Accuracy & Precision & Recall     \\ \midrule
		NSE                        & 70.0     & 76.5      & 76.5     & 90.4     & 84.5      & 79.6       \\
		DTEL                       & 86.2     & 88.2      & 88.2     & 86.3     & 73.4      & 71.4       \\
		Condor                     & 95.6     & 93.2      & \textbf{99.8}     & 95.4     & 91.1      & 90.8       \\
		MOOE                       & \textbf{97.1}     & \textbf{94.0}      & \textbf{99.8}     & \textbf{96.0}     & \textbf{93.4}      & \textbf{92.8}       \\
		\bottomrule
	\end{tabular}
	\caption{Comparisons on data with recurring concept drift.}
	\label{tb:comp2}
\end{table}

\subsubsection{Non-stationary Time Series with Recurring Concept Drift}
To verify the versatility of the proposed MOOE method, we conduct the comparisons on a special case of non-stationary time series, i.e., recurring concept drift, in which previous distributions may disappear and then re-appear in the future. We consider two real-world non-stationary time series with recurring concept drift, namely Email list and Spam filtering~\cite{RD:10}. The concepts are decided by the personal interests of users that change in a recurring manner. The results are summarized in~\cref{tb:comp2}, which show that MOOE exhibits an encouraging performance on the two datasets regarding all measures. Specifically, MOOE achieves $11.8 \%$ improvement in terms of accuracy. This is because the offline experts can learn the knowledge on the previous distributions, and the meta expert can reuse the knowledge when the distribution re-appear.

\begin{table}[h]
	\centering
	\begin{tabular}{ccccc}
		\toprule
		Approach      & NSE 	& DTEL & Condor & MOOE\\ \midrule
		Accuracy 		    & 64.9 & 58.7 & 80.1 & \bf{86.5}  \\ \bottomrule
	\end{tabular}
	\caption{Comparisons on data with increasing noise.}
	\label{tb:gn}
\end{table}

\subsubsection{Non-stationary Time Series with Increasing Levels of Noise}
To verify the robustness of the proposed MOOE method, we perform experiments on non-stationary time series with increasing levels of noise. Specifically, we adopt Covertype and gradually add Gaussian noise until the time series becomes completely random. The experiment results presented in~\cref{tb:gn} indicate that MOOE achieves the best prediction result. This is because MOOE can utilize the knowledge learned by the offline experts when the online expert is hard to learn knowledge from the noisy data.

\begin{figure}
	\centering
	\includegraphics[width=0.35\textwidth]{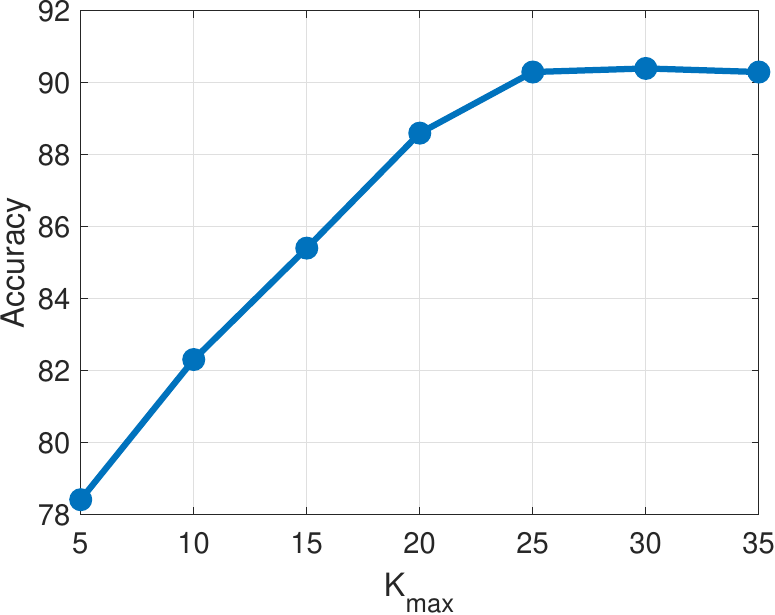}
	\caption{MOOE with different $K_{\text{max}}$.}
	\label{fig:para}
\end{figure}

\subsection{Effect of Maximal Number of Maintained Experts}
To verify the effect of the hyper-parameter $K_{\text{max}}$, we select its value from $\{5, 10, 15, 20, 25, 30, 35\}$ and perform experiments on Covertype. The experimental results presented in~\cref{fig:para} show that increasing $K_{\text{max}}$ can improve the classification performance and the performance stabilizes when $K_{\text{max}}$ is sufficiently large, e.g., $K_{\text{max}} = 25$. This is because a large number of maintained experts $K_{\text{max}}$ causes more knowledge can be stored and applied for the data in the online interval. Furthermore, if $K_{\text{max}}$ is sufficiently large, increasing its number cannot make MOOE obtain the new knowledge that the offline experts have not explored.

\section{Conclusion}
In this paper, we address a general and realistic scenario involving non-stationary time series, where several offline intervals with various distributions exist alongside an online interval. We propose MOOE, which employs a meta expert to integrate static offline experts, learned from previous offline intervals, with the dynamic online expert, updated in the online interval. We provide theoretical guarantees regarding parameter convergence, regret bounds, and generalization error bounds. Our theoretical results demonstrate that MOOE achieves the same generalization error bounds in both stationary and non-stationary cases, proving that leveraging knowledge from historical intervals is effective. Future work will explore other assumptions and techniques to overcome bottlenecks in the generalization bound.


\newpage

\section{Acknowledgments}
This work was supported by the Australian Research Council through the Linkage Grant (LP230201022), the Discovery Grant (DP240102050), and the Linkage Infrastructure, Equipment, and Facilities Grant (LE240100131).

\bibliography{ref}

\appendix

\section*{Proofs}
In this section, we present the proofs of all the theorems and lemmas. Our analysis follows some advanced techniques, including the self-bound property of smooth functions~\citep{SM:10}, the analysis of adaptive online optimization method with multiple learning rates~\citep{MG:16}, the connection between agnostic PAC learning and online convex optimization~\citep{OL:16}, empirical risk minimization for stochastic convex optimization~\citep{ERM:17}, and the bound of Rademacher complexity for any norm-regularized hypothesis class~\citep{LRC:18}.

\subsection{Proof of \cref{le:hy1}}
According to the property of strong convexity~\cite[Lemma 13.5.2]{ML:14}, we know $L_{\mathcal{\widetilde{S}}}^{\gamma}(\mathbf{w})$ is $\gamma$-strongly convex since $L_{\mathcal{S}}(\mathbf{w})$ is convex and $\frac{\gamma}{2} \Omega(\mathbf{w})$ is $\gamma$-strongly convex. Accordingly, we have
\begin{equation}
\begin{aligned}
\frac{\gamma}{2} \Omega \left(\mathbf{w}^K \right) & \leq L_{\mathcal{\widetilde{S}}}\left(\sum_{k = 1}^K \alpha_B^k \mathbf{w}_B^k \right) + \frac{\gamma}{2} \Omega\left(\sum_{k = 1}^K \alpha_B^k \mathbf{w}^k_B\right) \\
& - L_{\mathcal{\widetilde{S}}}\left(\mathbf{w}^K\right) - \frac{\gamma}{2} \Omega\left(\mathbf{w}^K\right) \\
\leq & \sum_{k = 1}^K \alpha_B^k L_{\mathcal{\widetilde{S}}}\left(\mathbf{w}_B^k\right) + 0 - 0 - \frac{\gamma}{2} \Omega\left(\mathbf{w}^K\right).\\
\end{aligned}
\end{equation}
Above, the first inequality is due to the strongly-convex property that $\frac{\gamma}{2} \Vert x - x^*\Vert_2^2 \leq f(x) - f(x^*)$~\citep[Lemma 13.5.3]{ML:14} and $\mathbf{w}^K$ is an empirical minimizer of $L_{\mathcal{\widetilde{S}}}^{\gamma}(\mathbf{w})$; the second inequality uses the condition that $L_{\mathcal{\widetilde{S}}}(\mathbf{w}^K) \geq 0$ as assumed.

\subsection{Proof of \cref{th:hy2}}
By using the fact that $\mathbf{w}^*$ minimizes $L_\mathcal{D}(\mathbf{w})$ over the domain $\mathcal{W}$, we have
\begin{equation}
\label{eq:hy2:1}
L_\mathcal{D}(\mathbf{w}^*) - L_\mathcal{D}(\mathbf{w}^K) \leq 0.
\end{equation}
According to the property of strong convexity~\cite[Lemma 13.5.2]{ML:14}, $L_\mathcal{D}(\mathbf{w}) + \frac{\gamma}{2} \Omega(\mathbf{w})$ is $\gamma$-strongly convex because the former term is convex and the last term is $\gamma$-strongly convex. Following the definition of strongly convex function, we have
\begin{equation}
\begin{aligned}
\label{eq:hy2:2}
& \frac{\gamma}{2} \left\Vert \mathbf{w}^K - \mathbf{w}^*\right\Vert_2^2 \\
\leq &  L_\mathcal{D}(\mathbf{w}^*) + \frac{\gamma}{2} \Omega(\mathbf{w}^*) - L_\mathcal{D}(\mathbf{w}^K) - \frac{\gamma}{2} \Omega(\mathbf{w}^K)  \\
+ &\left< \nabla L_\mathcal{D}(\mathbf{w}^K) + \gamma \left( \mathbf{w}^K - \sum_{k = 1}^K \alpha_B^k \mathbf{w}^k_B\right) , \mathbf{w}^K - \mathbf{w}^*\right>.\\
\end{aligned}
\end{equation}
To upper bound the last term above, we have
\begin{equation}
\begin{aligned}
\label{eq:hy2:3}
& \left< \nabla L_\mathcal{D}\left(\mathbf{w}^K\right) + \gamma \left( \mathbf{w}^K - \sum_{k = 1}^K \alpha_B^k \mathbf{w}^k_B\right) , \mathbf{w}^K - \mathbf{w}^*\right>\\
\leq & \left( \left\Vert \nabla L_\mathcal{D}(\mathbf{w}^K) \right\Vert + \gamma \left\Vert \mathbf{w}^K - \sum_{k = 1}^K \alpha_B^k \mathbf{w}^k_B \right\Vert \right) \left\Vert \mathbf{w}^K - \mathbf{w}^*\right\Vert\\
\leq &  \frac{4\left\Vert \nabla L_\mathcal{D}\left(\mathbf{w}^K\right) \right\Vert^2_2}{2 \gamma} + \frac{\gamma \left\Vert \mathbf{w}^K - \mathbf{w}^*\right\Vert^2_2}{2 \cdot 4} \\
& + \gamma \left( \frac{4 \left\Vert \mathbf{w}^K - \sum_{k = 1}^K \alpha_B^k \mathbf{w}^k_B \right\Vert_2^2}{2} + \frac{\left\Vert \mathbf{w}^K - \mathbf{w}^*\right\Vert^2_2}{2 \cdot 4} \right)\\
\end{aligned}
\end{equation}
where the first inequality uses the Cauchy-Schwarz inequality and the second inequality uses the Young's inequality $\left<a,b\right> \leq \frac{1}{2} \Vert a \Vert_2^2 + \frac{1}{2} \Vert b \Vert_2^2$.

Substituting Eqs. (\ref{eq:hy2:1}) and (\ref{eq:hy2:3}) into Eq.  (\ref{eq:hy2:2}), we have
\begin{equation}
\begin{aligned}
& \left\Vert \mathbf{w}^K - \mathbf{w}^*\right\Vert^2_2 \\
\leq & 2 \Omega\left(\mathbf{w}^*\right)  + \frac{8}{\gamma^2}\left\Vert \nabla L_\mathcal{D}\left(\mathbf{w}^K\right) \right\Vert^2_2 + 6 \Omega\left(\mathbf{w}^K\right) \\
\leq  & 2 \Omega\left(\mathbf{w}^*\right)  + \frac{32L}{\gamma^2} + \frac{6 \sum_{k = 1}^K \alpha_B^k L_{\mathcal{\widetilde{S}}}\left(\mathbf{w}_B^k\right)}{\gamma}
\end{aligned}
\end{equation}
where the second inequality is owing to Eq. (\ref{eq:func}) and \cref{le:hy1}.

\subsection{Proof of \cref{th:rg}}
As shown in Eq. (\ref{th:sp}), the analysis is divided into two parts. First, we show the $\text{Regret}_\text{ME}$ of the meta-expert: the difference between the total cost it has incurred and that of the best existing expert of $K$ experts. Then, we demonstrate the $\text{Regret}_\text{OE}$ of the online expert: the difference between the total cost it has incurred and that of the empirical minimizer. Based on the pervious study~\citep[Theorem 2.2]{PLG:06}, we define $W_j = \sum_{k = 1}^K \alpha_1^k e^{- \nu \sum_{t = 1}^j f_t(\mathbf{w}_t^k)}$
and the lower bound of the related quantities is
\begin{equation}
\begin{aligned}
\label{eq:rg:1}
\ln \frac{W_T}{W_0} = & \ln \left( \frac{\sum_{k = 1}^K \alpha_1^k e^{- \nu \sum_{t = 1}^T f_t(\mathbf{w}_t^k)}}{\sum_{k = 1}^K \alpha_1^k} \right) \\
= & \ln\left( \sum_{k = 1}^K \alpha_1^k e^{- \nu \sum_{t = 1}^T f_t(\mathbf{w}_t^k)}\right) - \ln\left( \sum_{k = 1}^K \alpha_1^k\right) \\
\geq & \ln \max_{k \in [K]} \left( \alpha_1^k e^{-\nu \sum_{t = 1}^T f_t(\mathbf{w}_t^k)}\right) - \ln\left( \sum_{k = 1}^K \alpha_1^k\right) \\
= & -\nu \min_{k \in [K]} \left(\sum_{t = 1}^T f_t(\mathbf{w}_t^k) + \frac{1}{\nu}\ln\frac{1}{\alpha_1^k} \right) - \ln\left( \sum_{k = 1}^K \alpha_1^k\right). \\
\end{aligned}
\end{equation}

On the other hand, for $j \in [T]$ and $k \in [K]$, we can use the updating rule of $\alpha_j^k$ defined in Eq. (\ref{eq:update:a}) to obtain
\begin{equation}
\begin{aligned}
\label{eq:rg:ho1}
\ln \frac{W_j}{W_{j - 1}} = &\log \left( \frac{\sum_{k = 1}^K \alpha_1^k e^{- \nu \sum_{t = 1}^j f_t(\mathbf{w}_t^k)}}{\sum_{k = 1}^K \alpha_1^k e^{- \nu \sum_{t = 1}^{j - 1} f_t(\mathbf{w}_t^k)}} \right) \\
= & \ln \left( \sum_{k = 1}^{K} \frac{\alpha_1^k e^{- \nu \sum_{t = 1}^{j - 1} f_t(\mathbf{w}_t^k)}}{\sum_{k = 1}^K \alpha_1^k e^{- \nu \sum_{t = 1}^{j - 1} f_t(\mathbf{w}_t^k)}} e^{- \nu f_j(\mathbf{w}_j^k)}\right) \\
=  & \ln \left( \sum_{k = 1}^{K} \alpha_j^k e^{- \nu f_j(\mathbf{w}_j^k)}\right). \\
\end{aligned}
\end{equation}

To bound the above result further, we require the following Hoeffding's Lemma.
\begin{lemma}\citep[Lemma 2.2]{PLG:06}
\label{le:Hoeffding}
Let $X$ be a random variable with $a \leq X \leq b$, then for any $s \in \mathbb{R}$,
\begin{align*}
\ln E [e^{s X}] \leq s E[X] + \frac{s^2 (b - a)^2}{8}.
\end{align*}
\end{lemma}
Recalling Assumption~\ref{as:4} that $0 \leq f_t(\cdot) \leq 1$ and combining that with Lemma~\ref{le:Hoeffding}, we have
\begin{equation}
\begin{aligned}
\label{eq:rg:ho2}
\ln \left( \sum_{k = 1}^{K} \alpha_j^k e^{- \nu f_j(\mathbf{w}_j^k)}\right) \leq & - \nu \sum_{k = 1}^K \alpha_t^k f_t(\mathbf{w}_t^k) + \frac{\nu^2 \left(1 - 0\right)^2}{8} \\
\leq & - \nu f_t(\mathbf{w}_t) + \frac{\nu^2}{8}\\
\end{aligned}
\end{equation}
where the last inequality is owing to the Jensen's inequality.

Substituting Eqs. (\ref{eq:rg:ho2}) into (\ref{eq:rg:ho1}) and accumulating the result from $t = 1$ to $t = T$, we have
\begin{equation}
\begin{aligned}
\label{eq:rg:2}
\sum_{j = 1}^T \ln \frac{W_j}{W_{j - 1}} = \ln \frac{W_T}{W_0} =  - \nu \sum_{t = 1}^T f_t(\mathbf{w}_t) + \frac{T\nu^2}{8}.
\end{aligned}
\end{equation}
By combining Eq. (\ref{eq:rg:1}) with Eq. (\ref{eq:rg:2}), we have
\begin{equation}
\begin{aligned}
& \sum_{t = 1}^T f_t(\mathbf{w}_t) - \min_{k \in [K]} \left(\sum_{t = 1}^T f_t(\mathbf{w}_t^k) + \frac{1}{\nu}\ln\frac{1}{\alpha_1^k} \right) \\
\leq &\frac{T \nu}{8} + \ln\left( \sum_{k = 1}^K \alpha_1^k\right)\\
\end{aligned}
\end{equation}
which implies
\begin{equation}
\begin{aligned}
\label{eq:rg:3}
 & \sum_{t = 1}^T f_t(\mathbf{w}_t) - \min_{k \in [K]} \sum_{t = 1}^T f_t(\mathbf{w}_t^k) \\
 \leq & \frac{T \nu}{8} + \ln\left( \sum_{k = 1}^K \alpha_1^k\right) + \max_{k \in [K]} \frac{1}{\nu}\ln\frac{1}{\alpha_1^k}.
\end{aligned}
\end{equation}
With Eq. (\ref{eq:in}), we have
\begin{equation}
\begin{aligned}
\label{eq:rg:as}
& \sum_{k = 1}^K \alpha_1^k = \frac{K + 1}{K}\left[ \left( \frac{1}{K } - \frac{1}{K + 1}\right)  + ,\ldots,  + \left( \frac{1}{1} - \frac{1}{2}\right)\right] \\
= & \frac{K + 1}{K} \cdot \frac{K }{K + 1 } = 1,
\end{aligned}
\end{equation}
and for any $2 \leq k \leq K$, we have
\begin{equation}
\begin{aligned}
\frac{K}{K + 1} \left( \alpha_1^{k - 1} - \alpha_1^k \right) & \geq (\frac{1}{K - k} - \frac{1}{K + 1 - k}) \\
& - (\frac{1}{K + 1 - k} - \frac{1}{K + 2 - k}) \geq 0,
\end{aligned}
\end{equation}
so $\{\alpha_1^{k}\}$ is in the ascending order and
\begin{equation}
\begin{aligned}
\label{eq:rg:ab}
\max_{k \in [K]}\ln\frac{1}{\alpha_1^k} = \ln\frac{1}{\alpha_1^1} = 2 \ln K.
\end{aligned}
\end{equation}
Substitute Eqs. (\ref{eq:rg:as}) and (\ref{eq:rg:ab}) into Eq. (\ref{eq:rg:3}), we have
\begin{align}
\label{eg:rg:mee}
\frac{T \nu}{8} + \frac{2}{\nu}\ln K,
\end{align}
and minimize this function toward $\nu$ to obtain
\begin{align}
\label{eq:rg:v}
\nu = 4 \sqrt{\frac{\ln K}{T}}.
\end{align}
Substitute Eqs.~(\ref{eq:in}) and~(\ref{eq:rg:v}) into Eq.~(\ref{eg:rg:mee}), we have
\begin{equation}
\begin{aligned}
\label{eq:rg:me}
\text{Regret}_\text{ME} \leq \sqrt{T \ln K}.
\end{aligned}
\end{equation}
We apply the standard OSG to optimize the online expert $\mathbf{w}^K_t$. Following the previous result~\citep[Theorem 3.1]{OL:16}, setting the step size $\eta_t = \frac{D}{\sqrt{\beta t}}$, we have
\begin{equation}
\begin{aligned}
\label{eq:rg:oe}
\text{Regret}_\text{OE} \leq 6D\sqrt{T \beta}.
\end{aligned}
\end{equation}
We obtain $\text{Regret}_{\text{CO$_2$}}$ by combining Eq.~(\ref{eq:rg:me}) with Eq.~(\ref{eq:rg:oe}).
According to Eq. (\ref{th:sp}) and the requirement that CO$_2$ should surpass its online expert, we obtain the upper bound of $K$ by solving the following inequality
\begin{align*}
\sqrt{T \ln K} + \text{Regret}_\text{KE} \leq 6D\sqrt{T \beta}.
\end{align*}

\subsection{Proof of \cref{le:erm:rg}}
To proceed, we introduce the following norm concentration inequality.
\begin{lemma}\cite[Proposition 1]{CM:09}
\label{eq:cm:e}
Let $\xi$ be a random variable on $(Z,\rho)$ with values in a Hilbert space and be randomly drawn according to $\rho$ satisfying $\Vert \xi \Vert \leq M \leq \infty$. Then, for any $0 < \delta < 1$, with a probability at least $1 - \delta$,
\begin{align*}
\left\Vert \frac{1}{m}\sum_{i = 1}^m \left( \xi_i - \mathbb{E}[\xi_i]\right) \right\Vert \leq \frac{4M}{\sqrt{m}} \log \frac{2}{\delta}.
\end{align*}
\end{lemma}
Using Lemma~\ref{eq:cm:e} with the results $\Vert \frac{1}{T} \sum_{t = 1}^T f_t(\mathbf{w}_t) \Vert \leq 1$ and $\Vert \frac{1}{T} \sum_{t = 1}^T f_t(\mathbf{w}^*) \Vert \leq 1$ implied from Assumption~\ref{as:4}, with probability at least $1 - \delta$, we have
\begin{equation}
\begin{aligned}
\label{eq:all:2}
\frac{1}{T} \sum_{t = 1}^T L_\mathcal{D}(\mathbf{w}_t) & \leq \frac{1}{T} \sum_{t = 1}^T f_t(\mathbf{w}_t) + \frac{4 \log(2 / \delta)}{\sqrt{T}}, \frac{1}{T} \sum_{t = 1}^T f_t(\widehat{\mathbf{w}}) \\
& \leq \frac{1}{T} \sum_{t = 1}^T L_\mathcal{D}(\widehat{\mathbf{w}}) + \frac{4 \log(2 / \delta)}{\sqrt{T}}. \\
\end{aligned}
\end{equation}
Putting the two results in Eq. (\ref{eq:all:2}) together, with probability at least $1 - \delta$, we have
\begin{equation}
\begin{aligned}
\label{eq:all:3}
& \frac{1}{T} \sum_{t = 1}^T L_\mathcal{D}(\mathbf{w}_t) - L_\mathcal{D}(\widehat{\mathbf{w}}) \\
\leq & \frac{1}{T} \left( \sum_{t = 1}^T f_t(\mathbf{w}_t) - \sum_{t = 1}^T f_t(\widehat{\mathbf{w}})\right) + \frac{4 \log(4 / \delta)}{\sqrt{T}} \\
\leq & \frac{\sqrt{\ln K} + 6D\sqrt{\beta}}{\sqrt{T}} + \frac{4 \log(4 / \delta)}{\sqrt{T}}\\
\end{aligned}
\end{equation}
where the first inequality is owing to the Hoeffding's inequality~\cite[Theorem 2.8]{CMB:13} and the second inequality follows \cref{th:rg}.

\subsection{Proof of Lemma~\ref{le:erm:di}}
Our analysis is based on the techniques used in~\citep{ERM:17}. For simplicity, we assume $\dot{b}_t = \nabla f_t(\mathbf{w}^*)$ and $\ddot{b}_t = \nabla f_t(\widehat{\mathbf{w}}) -  \nabla f_t(\mathbf{w}^*)$, so we know
\begin{equation}
\begin{aligned}
& \nabla L_\mathcal{S}(\mathbf{w}^*)  = \frac{1}{T} \sum_{t = 1}^T \dot{b}_t ,\\
& \nabla L_\mathcal{D}(\mathbf{w}^*) = \mathbb{E} [\dot{b}], \\
& \nabla L_\mathcal{S}(\widehat{\mathbf{w}}) - \nabla L_\mathcal{S}(\mathbf{w}^*) = \frac{1}{T} \sum_{t = 1}^T \ddot{b}_t ,\\
& \nabla L_\mathcal{D}(\widehat{\mathbf{w}}) - \nabla L_\mathcal{D}(\mathbf{w}^*) = \mathbb{E} [\ddot{b}].\\
\end{aligned}
\end{equation}
By the Karush-Kuhn-Tucker (KKT)~\citep[Theorem 2.2]{OL:16} condition for convex function $L_\mathcal{S}(\cdot)$ and $L_\mathcal{D}(\cdot)$, we have
\begin{equation}
\begin{aligned}
& \left<\nabla L_\mathcal{S}(\widehat{\mathbf{w}}), \mathbf{w} - \widehat{\mathbf{w}}\right> \geq 0, \left<\nabla L_\mathcal{D}(\mathbf{w}^*), \mathbf{w} - \mathbf{w}^*\right> \geq 0, \forall \mathbf{w} \in \mathcal{W}. \\
\end{aligned}
\end{equation}
We first upper bound the excess risk $L_\mathcal{D}(\widehat{\mathbf{w}}) - L_\mathcal{D}(\mathbf{w}^*)$ by two terms with the same form and then further derive the upper bounds of the two terms by similar methods.
\begin{equation}
\begin{aligned}
\label{eq:erm1:1}
& L_\mathcal{D}(\widehat{\mathbf{w}}) - L_\mathcal{D}(\mathbf{w}^*) \\
\leq & \left< \nabla L_\mathcal{D}(\widehat{\mathbf{w}}) , \widehat{\mathbf{w}} - \mathbf{w}^*\right> =  \left< \mathbb{E} [\ddot{b}] + \mathbb{E} [\dot{b}] , \widehat{\mathbf{w}} - \mathbf{w}^*\right> \\
= & \left< \mathbb{E} [\ddot{b}] - \frac{1}{T} \sum_{t = 1}^T \ddot{b}_t + \mathbb{E} [\dot{b}]  + \frac{1}{T} \sum_{t = 1}^T \ddot{b}_t , \widehat{\mathbf{w}} - \mathbf{w}^*\right> \\
\leq & \left< \mathbb{E} [\ddot{b}] - \frac{1}{T} \sum_{t = 1}^T \ddot{b}_t + \mathbb{E} [\dot{b}]  + \frac{1}{T} \sum_{t = 1}^T \dot{b}_t , \widehat{\mathbf{w}} - \mathbf{w}^*\right> \\
\leq & \underbrace{ \left \Vert \mathbb{E} [\ddot{b}] - \frac{1}{T} \sum_{t = 1}^T \ddot{b}_t \right\Vert \left\Vert \widehat{\mathbf{w}} - \mathbf{w}^* \right\Vert }_{:=B_1} \\
& + \underbrace{\left\Vert \mathbb{E} [\dot{b}] - \frac{1}{T} \sum_{t = 1}^T \dot{b}_t \right\Vert \left\Vert \widehat{\mathbf{w}} - \mathbf{w}^* \right\Vert}_{:B_2}.
\end{aligned}
\end{equation}
In the above, the first inequality is owing to the convexity of $L_{\mathcal{D}}(\cdot)$ over the domain $\mathcal{W}$, the second inequality applies Eq. (\ref{eq:erm1:1}) for convex function $L_\mathcal{S}(\cdot)$ with respect to $\mathbf{w}^*$: $\left<\nabla L_\mathcal{S}(\widehat{\mathbf{w}}), \widehat{\mathbf{w}} - \mathbf{w}^*\right> \leq 0$ and uses $\mathbb{E} [\dot{b}]  + \nabla L_\mathcal{S}(\widehat{\mathbf{w}}) - \frac{1}{T} \sum_{t = 1}^T \dot{b}_t$; the third inequality uses the Cauchy-Schwarz inequality.

Note that $B_1$ and $B_2$ have the same structure. To bound the variance terms in $B_1$ and $B_2$, we introduce the following norm concentration inequality in a Hilbert space.
\begin{lemma}\citep[Lemma 2]{CM:07}
\label{le:cm}
Let $\xi$ be a random variable on $(Z,\rho)$ with values in a Hilbert space, assume $\Vert \xi \Vert \leq M \leq \infty$ almost surely holds, denote $\sigma^2(\xi) = \mathbb{E}(\Vert \xi \Vert^2)$, and let ${\xi_i}_{i = 1}^m$ be independent random drawers of $\rho$, for any $0 \leq \delta \leq 1$, with confidence $1 - \delta$,
\begin{align*}
\left\Vert \frac{1}{m}\sum_{i = 1}^m \left( \xi_i - \mathbb{E}[\xi_i]\right) \right\Vert \leq \frac{2 M \log(2/\delta)}{m} + \sqrt{\frac{2 \sigma^2(\xi) \log(2/\delta)}{m}}.
\end{align*}
\end{lemma}
To use this lemma to upper bound $B_1$ and $B_2$, we need the bounds for $\Vert \dot{b}_t \Vert$, $\mathbb{E}\Vert \dot{b} \Vert^2_2$, $\Vert \ddot{b}_t \Vert$, $\mathbb{E}\Vert \ddot{b} \Vert^2_2$. From Eq. (\ref{eq:func}), we have
\begin{equation}
\begin{aligned}
\label{eq:erm1:2}
\Vert \dot{b}_t \Vert \leq 2 \sqrt{\beta}, \mathbb{E}\Vert \dot{b} \Vert^2_2 \leq 4 \beta.\\
\end{aligned}
\end{equation}
With Assumption~\ref{as:5}, we have
\begin{align*}
\Vert \ddot{b}_t \Vert \leq \beta \Vert \widehat{\mathbf{w}} - \mathbf{w}^* \Vert.
\end{align*}
Based on that, by using the properties of smooth function~\citep[Theorem 2.1.5]{Nes:04}, we have
\begin{align}
\label{eq:erm1:3}
\Vert \ddot{b}_t \Vert_2^2 \leq 2 \beta \left(f_t(\widehat{\mathbf{w}}) - f_t(\mathbf{w}^*) - \left<\nabla f_t(\widehat{\mathbf{w}}) , \widehat{\mathbf{w}} - \mathbf{w}^* \right> \right).
\end{align}
Taking the expectation on both sides, we have
\begin{equation}
\begin{aligned}
\label{eq:erm1:4}
\mathbb{E} \Vert \ddot{b} \Vert_2^2 & \leq 2 \beta \left(L_{\mathcal{D}}(\widehat{\mathbf{w}}) - L_{\mathcal{D}}(\mathbf{w}^*) - \left<\nabla L_{\mathcal{D}}(\widehat{\mathbf{w}}) , \widehat{\mathbf{w}} - \mathbf{w}^* \right> \right) \\
\leq & 2 \beta \left(L_{\mathcal{D}}(\widehat{\mathbf{w}}) - L_{\mathcal{D}}(\mathbf{w}^*) \right),\\
\end{aligned}
\end{equation}
where the last inequality applies Eq. (\ref{eq:erm1:1}) to the convex function $L_{\mathcal{D}}(\cdot)$ with respect to $\widehat{\mathbf{w}}$.

Based on Lemma~\ref{le:cm}, we establish the uniform convergence of $\frac{1}{T} \sum_{t = 1}^T \dot{b}_t$ to $\mathbb{E} [\dot{b}]$ and $\frac{1}{T} \sum_{t = 1}^T \ddot{b}_t$ to $\mathbb{E} [\ddot{b}]$. By using Eqs. (\ref{eq:erm1:3}) and (\ref{eq:erm1:4}) with Lemma~\ref{le:cm}, with probability at least $1 - \delta$, we have
\begin{equation}
\begin{aligned}
\label{eq:erm1:5}
B_1 \leq & \frac{2\beta \Vert \widehat{\mathbf{w}} - \mathbf{w}^* \Vert_2^2 \log(2 / \delta) }{T} \\
& + 2 \Vert \widehat{\mathbf{w}} - \mathbf{w}^* \Vert \sqrt{\frac{\beta L_{\mathcal{D}}(\widehat{\mathbf{w}}) - L_{\mathcal{D}}(\mathbf{w}^*) \log(2 / \delta) }{T}}\\
\leq & \frac{3\beta \Vert \widehat{\mathbf{w}} - \mathbf{w}^* \Vert_2^2 \log(2 / \delta) }{T} + \frac{L_{\mathcal{D}}(\widehat{\mathbf{w}}) - L_{\mathcal{D}}(\mathbf{w}^*)}{2}  \\
\leq & \frac{12\beta R^2 \log(2 / \delta) }{T} + \frac{L_{\mathcal{D}}(\widehat{\mathbf{w}}) - L_{\mathcal{D}}(\mathbf{w}^*)}{2},\\
\end{aligned}
\end{equation}
where the second inequality uses the Young's inequality and the third inequality is owing to that Assumption~\ref{as:2} implies $\Vert \widehat{\mathbf{w}} - \mathbf{w}^* \Vert \leq 2R$. By using the same method as above with Eq. (\ref{eq:erm1:2}), with probability at least $1 - \delta$, we have
\begin{equation}
\begin{aligned}
\label{eq:erm1:6}
B_2 \leq \frac{4R \sqrt{\beta} \log(2/\delta)}{T} + 4R \sqrt{\frac{2 \beta \log(2/\delta)}{T}}.
\end{aligned}
\end{equation}
We complete the proof by substituting Eqs. (\ref{eq:erm1:5}) and (\ref{eq:erm1:6}) into Eq. (\ref{eq:erm1:1}).

\subsection{Proof of \cref{le:erm:dd}}
Based on the divide-and-conquer idea, we split our target into three more comfortable summands and derive their bounds respectively,
\begin{equation}
\begin{aligned}
\label{eq:erm2:1}
& L_\mathcal{D}(\widehat{\mathbf{w}}) - L_\mathcal{D}(\mathbf{w}^*)
=  \left( L_\mathcal{D}(\widehat{\mathbf{w}}) - L_\mathcal{S}(\widehat{\mathbf{w}}) \right) \\
& + \left( L_\mathcal{S}(\widehat{\mathbf{w}}) - L_\mathcal{S}(\mathbf{w}^*) \right) + \left( L_\mathcal{S}(\mathbf{w}^*) - L_\mathcal{D}(\mathbf{w}^*) \right).
\end{aligned}
\end{equation}
Recall that $l$ is a nonnegative function. To bound the generation error $L_\mathcal{D}(\widehat{\mathbf{w}}) - L_\mathcal{S}(\widehat{\mathbf{w}})$ by the Rademacher complexity for this nonnegative function, we need the following two lemmas.
\begin{lemma}\citep[Theorem 26.5]{ML:14}
\label{le:data_bound}
Assume that the loss function $l(h,z)$ is bounded by $b$ for all $z$ and $h \in \mathcal{H}$, $S$ is a $m$-samples data set. With probability of at least $1 - \delta$, for all $h \in \mathcal{H}$,
\begin{align*}
L_{\mathcal{D}}(h) - L_{\mathcal{S}}(h) \leq 2 \mathcal{R}(l\circ\mathcal{H}) + b \sqrt{\frac{2 \ln(2 / \delta)}{m}}.
\end{align*}
\end{lemma}
\begin{lemma}\citep[Lemma 2.2]{SM:10}
\label{le:contraction}
For a nonnegative $\beta$-smooth loss $l$ bounded by $b$ and any function class $\mathcal{H}$, the Rademacher Complexity on a $m$-sample data set is
\begin{align*}
\mathcal{R}(l \circ \mathcal{H})  \leq 21\sqrt{6 \beta b} \log^{\frac{3}{2}}(64m) \mathcal{R}(\mathcal{H}).
\end{align*}
\end{lemma}
With Assumption~\ref{as:4}, we know $\vert l( \left<\mathbf{w},\mathbf{x} \right>, y) \vert \leq 1$. By combing Lemmas~\ref{le:data_bound} and ~\ref{le:contraction} under the condition that $\widehat{\mathbf{w}} \in \mathcal{W}$, with probability at least $1 - \delta$, we have
\begin{equation}
\begin{aligned}
\label{eq:erm2:5}
& L_\mathcal{D}(\widehat{\mathbf{w}}) - L_\mathcal{S}(\widehat{\mathbf{w}})
\leq  2 \mathcal{R}_{\mathcal{D}}(l\circ\mathcal{W}) + \sqrt{\frac{2 \log(2 / \delta) }{T}} \\
\leq & 42\sqrt{6 \beta} \log^{\frac{3}{2}}(64T) \mathcal{R}_{\mathcal{D}}(\mathcal{W})  + \sqrt{\frac{2 \log(2 / \delta) }{T}}.
\end{aligned}
\end{equation}
Note that the above result ignores the specificity of $\widehat{\mathbf{w}}$. We utilize the specificity of $\widehat{\mathbf{w}}$ as well as $\mathbf{w}^*$ to bound the next two summands.

Recall that $\widehat{\mathbf{w}}$ is an empirical minimizer of $L_\mathcal{S}(\mathbf{w})$ over the domain $\mathcal{W}$, we have
\begin{align}
\label{eq:erm2:2}
L_\mathcal{S}(\widehat{\mathbf{w}}) - L_\mathcal{S}(\mathbf{w}^*) \leq 0.
\end{align}

Because $\mathbf{w}^*$ is independent of the data set $\mathcal{S}$ and $f_1(\mathbf{w}^*), \ldots ,f_T(\mathbf{w}^*)$ is a sequence of i.i.d. random variables, we have $\mathbb{E}_{\mathcal{S}'} [L_{\mathcal{S}'}(\mathbf{w}^*) ] = L_\mathcal{D}(\mathbf{w}^*)$. With Assumption~\ref{as:4}, we know that $\mathbb{P}[0 \leq f_t(\mathbf{w}^*) \leq 1] = 1$ for every $t \in [T]$. The Hoeffding's inequality~\citep[Theorem 2.8]{CMB:13} implies that with probability at least $1 - \delta$, we have
\begin{align}
\label{eq:erm2:3}
L_\mathcal{S}(\mathbf{w}^*) - L_\mathcal{D}(\mathbf{w}^*)\leq \sqrt{\frac{\log(2 / \delta)}{2 T}}.
\end{align}
We complete the proof by substituting Eqs. (\ref{eq:erm2:2}), (\ref{eq:erm2:3}) and (\ref{eq:erm2:5}) into Eq. (\ref{eq:erm2:1}).

\subsection{Proof of \cref{th:all}}
To proceed, we introduce the following lemma.
\begin{lemma}
\label{le:rademacher}
Let $\mathcal{S}$ be a set of i.i.d. samples from the distribution $\mathcal{D}$, the Rademacher complexity $\mathcal{R}$ of hypothesis class $\mathcal{H}$ w.r.t. distribution $\mathcal{D}$ at the online interval is bounded as
\begin{equation*}
\begin{aligned}
\mathcal{R}_\mathcal{D}(\mathcal{H}) \leq R\sqrt{\frac{1}{T}  \sum_{i = 1}^{\infty} \left( D^2 \wedge \frac{e \lambda_i}{T} \right)} +  \frac{DR\sqrt{e}}{T}.
\end{aligned}
\end{equation*}
\end{lemma}
We complete the proof by substituting the results of Lemmas~\ref{le:erm:rg},~\ref{le:erm:di},~\ref{le:rademacher} and~\ref{le:erm:dd} into the excess risk bound framework in Eq.~(\ref{eq:all:1}).

\subsection{Proof of Lemma~\ref{le:rademacher}}
By using the definition of the Rademacher complexity~\citep{RC:02}, we have
\begin{align*}
\mathcal{R}_{\mathcal{D}}(\mathcal{W}) = \mathbb{E}_{\mathbf{S},\mathbf{\sigma}} \left[ \sup_{\mathbf{w} \in \mathcal{W}} \frac{1}{T} \sum_{t = 1}^T \sigma_t \left< \mathbf{w},\mathbf{x}_t\right> \right].
\end{align*}

Let $\Gamma_i = \left<\sum_{t = 1}^T \sigma_t\mathbf{x}_t, \mathbf{u}_i\right>/T$ for any $i \geq 1$ and $\theta \geq i$, by using advanced techniques of the Rademacher complexities~\citep{LRC:18}, we have
\begin{align*}
& \mathbb{E}_{\mathbf{S},\mathbf{\sigma}} \left[ \sup_{\mathbf{w} \in \mathcal{W}}  \sigma_t \left< \mathbf{w},\frac{1}{T} \sum_{t = 1}^T \mathbf{x}_t\right> \right] \\
= & \mathbb{E}_{\mathbf{S},\mathbf{\sigma}} \left[ \sup_{\mathbf{w} \in \mathcal{W}}  \sigma_t \left< \mathbf{w}, \sum_{i = 1}^\infty \left< \frac{1}{T} \sum_{t = 1}^T \mathbf{x}_t, \mathbf{u}_i\right>\mathbf{u}_i \right> \right]\\
\leq & \mathbb{E}_{\mathbf{S},\mathbf{\sigma}} \left[ \sup_{\mathbf{w} \in \mathcal{W}} \left< \mathbf{w}, \sum_{i = 1}^\theta \Gamma_i \mathbf{u}_i \right> \right] \\
& + \mathbb{E}_{\mathbf{S},\mathbf{\sigma}} \left[ \sup_{\mathbf{w} \in \mathcal{W}}   \left< \mathbf{w}, \sum_{i = \theta + 1}^\infty \Gamma_i\mathbf{u}_i \right> \right]\\
\end{align*}
The above inequality is owing to the Jensen's inequality. Based on the Cauchy-Schwarz inequality, we have the following upper bounds for the last two terms
\begin{equation*}
\begin{aligned}
& \mathbb{E}_{\mathbf{S},\mathbf{\sigma}} \left[ \sup_{\mathbf{w} \in \mathcal{W}} \left< \mathbf{w}, \sum_{i = 1}^\theta \Gamma_i \mathbf{u}_i \right> \right] \\
= & \mathbb{E}_{\mathbf{S},\mathbf{\sigma}} \left[ \sup_{\mathbf{w} \in \mathcal{W}} \left< \sum_{i = 1}^\theta \sqrt{\lambda_i} \left<\mathbf{w}, \mathbf{u}_i\right>\mathbf{u}_i, \sum_{i = 1}^\theta \frac{1}{\sqrt{\lambda_i}} \Gamma_i\mathbf{u}_i \right> \right]\\
\leq & \mathbb{E}_{\mathbf{S},\mathbf{\sigma}} \left[ \sup_{\mathbf{w} \in \mathcal{W}} \left \Vert \sum_{i = 1}^\theta \sqrt{\lambda_i} \left<\mathbf{w}, \mathbf{u}_i\right>\mathbf{u}_i \right \Vert_2  \left\Vert \sum_{i = 1}^\theta \frac{1}{\sqrt{\lambda_i}} \Gamma_i\mathbf{u}_i \right\Vert_2  \right] \\
=  & \underbrace{\sup_{\mathbf{w} \in \mathcal{W}} \sqrt{ \sum_{i = 1}^\theta \lambda_i \left<\mathbf{w}, \mathbf{u}_i\right>^2}}_{:= U_1} \underbrace{\mathbb{E}_{\mathbf{S},\mathbf{\sigma}}\sqrt{ \sum_{i = 1}^\theta \frac{1}{\lambda_i} \Gamma_i^2 }}_{:= U_2},
\end{aligned}
\end{equation*}
and
\begin{equation*}
\begin{aligned}
& \mathbb{E}_{\mathbf{S},\mathbf{\sigma}} \left[ \sup_{\mathbf{w} \in \mathcal{W}}   \left< \mathbf{w}, \sum_{i = \theta + 1}^\infty \Gamma_i\mathbf{u}_i \right> \right] \\
\leq & \underbrace{ \sup_{\mathbf{w} \in \mathcal{W}} \left\Vert \mathbf{w} \right\Vert \mathbb{E}_{\mathbf{S},\mathbf{\sigma}} \left[   \left\Vert \sum_{i = \theta + 1}^\infty \Gamma_i\mathbf{u}_i \right\Vert \right]}_{:= U_3}.\\
\end{aligned}
\end{equation*}

\subsubsection{Bounding $U_1$:} Enlarging $U_1$ by replacing $\theta$ with $\infty$, we have the following upper bound
\begin{align*}
U_1 \leq & \sup_{\mathbf{w} \in \mathcal{W}} \sqrt{ \sum_{i = 1}^\infty \lambda_i \left<\mathbf{w}, \mathbf{u}_i\right>\left<\mathbf{w}, \mathbf{u}_i\right>} \\
= & \sup_{\mathbf{w} \in \mathcal{W}} \sqrt{\mathbb{E} \left<\mathbf{w} \mathbf{w}^T, [\mathbf{x} \mathbf{x}^T]\right>} \leq DR,
\end{align*}
where the last inequality uses Assumptions~\ref{as:2} and ~\ref{as:3}.

\subsubsection{Bounding $U_2$:} By using the Jensen's inequality, we have
\begin{align*}
U_2 \leq & \sqrt{ \sum_{i = 1}^\theta \mathbb{E}_{\mathbf{S},\mathbf{\sigma}}\frac{1}{\lambda_i} \left< \frac{1}{T} \sum_{t = 1}^T  \sigma_t \mathbf{x}_t, \mathbf{u}_i\right>^2 } \\
=  & \sqrt{ \sum_{i = 1}^\theta \mathbb{E}_{\mathbf{S},\mathbf{\sigma}}\frac{1}{\lambda_i T^2} \sum_{t = 1}^T \sum_{t' = 1}^T\sigma_t\sigma_t'\left< \mathbf{x}_t, \mathbf{u}_i\right>\left< \mathbf{x}_t', \mathbf{u}_i\right> }\\
= & \sqrt{\sum_{i = 1}^\theta \frac{1}{\lambda_i T} \left< \frac{1}{T}\sum_{t = 1}^T \mathbb{E}_{\mathbf{S}}\left[\mathbf{x}_t\mathbf{x}_t^T\right],\mathbf{u}_i\mathbf{u}_i^T\right> } \\
= & \sqrt{\sum_{i = 1}^\theta \frac{1}{\lambda_i T} \left< \sum_{j = 1}^\infty \lambda_j \mathbf{u}_j\mathbf{u}_j^T,\mathbf{u}_i\mathbf{u}_i^T\right> } \\
=  &\sqrt{\sum_{i = 1}^\theta \frac{1}{\lambda_i T} \lambda_i} = \sqrt{\frac{\theta}{T}}.
\end{align*}

\subsubsection{Bounding $U_3$:}
Recall the definition of $\Gamma_i$, we have
\begin{equation}
\begin{aligned}
\label{eq:u3:1}
U_3 \leq & R \mathbb{E}_{\mathbf{S},\mathbf{\sigma}} \sqrt{ \left\Vert \sum_{i = \theta + 1}^\infty \left< \frac{1}{T} \sum_{t = 1}^T \sigma_t \mathbf{x}_t, \mathbf{u}_i\right>\mathbf{u}_i \right\Vert^2_2 },\\
\end{aligned}
\end{equation}
where the inequality is owing to our assumption $\sup_{\mathbf{w} \in \mathcal{W}} \left\Vert \mathbf{w} \right\Vert \leq R$. To derive its bound further, we introduce the Khintchine-Kahane inequality.
\begin{lemma}\citep{KKI:12}
\label{le:KKI}
Let $\mathcal{H}$ be an inner-product space with the induced norm $\Vert \cdot \Vert_{\mathcal{H}}$, $v_1, \ldots ,v_n \in \mathcal{H}$ and $\sigma_1, \ldots, \sigma_n$ i.i.d. Rademacher random variables. Then, for any $p \geq 1$, we have
\begin{align*}
\mathbb{E} \Vert \sum_{i = 1}^n \sigma_i v_i\Vert_{\mathcal{H}}^p \leq \left( c \sum_{i = 1}^n \Vert v_i\Vert_{\mathcal{H}}^2 \right)^{\frac{p}{2}}
\end{align*}
where $c:= \max{1,p-1}$. The inequality also holds for $p$ in place of $c$.
\end{lemma}
With Lemma~\ref{le:KKI}, we have
\begin{equation}
\begin{aligned}
\label{eq:u3:2}
& \mathbb{E}_{\mathbf{S},\mathbf{\sigma}} \sqrt{ \left\Vert \sum_{i = \theta + 1}^\infty \left< \frac{1}{T} \sum_{t = 1}^T \sigma_t \mathbf{x}_t, \mathbf{u}_i\right>\mathbf{u}_i \right\Vert^2_2 } \\
\leq & \frac{1}{\sqrt{T}} \mathbb{E}_{\mathbf{S}} \sqrt{ \frac{1}{T} \sum_{t = 1}^T\left\Vert \sum_{i = \theta + 1}^\infty \left<\mathbf{x}_t, \mathbf{u}_i\right>\mathbf{u}_i \right\Vert^2_2 }  \\
= & \frac{1}{\sqrt{T}} \mathbb{E}_{\mathbf{S}} \sqrt{ \frac{1}{T} \sum_{t = 1}^T \sum_{i = \theta + 1}^\infty \left<\mathbf{x}_t, \mathbf{u}_i\right>^2 }.\\
\end{aligned}
\end{equation}

Although we can bound Eq. (\ref{eq:u3:2}) further according to the following result obtained from Assumption~\ref{as:3}
\begin{equation}
\begin{aligned}
\label{eq:u3:xx}
\sum_{i = \theta + 1}^\infty \left<\mathbf{x}_t, \mathbf{u}_i\right>^2 =  \left<\sum_{i = \theta + 1}^\infty \left<\mathbf{x}_t, \mathbf{u}_i\right> \mathbf{u}_i, \mathbf{x}_t\right> \leq \left<\mathbf{x}_i, \mathbf{x}_i\right> = D^2,
\end{aligned}
\end{equation}
we pursue a tighter bound by introducing the Rosenthal-Young inequality.
\begin{lemma}\citep[Lemma 3]{RYI:12}
\label{le:RYI}
Let the independent nonnegative random variables $X_1, \ldots ,X_n$ satisfy $X_i \leq B \leq +\infty$ almost surely for all $i = 1, \ldots ,n$, if $q \geq \frac{1}{2}$, $c_q := (2 q e)^q$, then the following holds
\begin{align*}
\mathbb{E} \left( \frac{1}{n} \sum_{i = 1}^n X_i\right)^q \leq c_q \left[ \left( \frac{B}{n}\right)^q + \left( \frac{1}{n} \sum_{i = 1}^n \mathbb{E} X_i\right)^q \right].
\end{align*}
\end{lemma}

By combining Eq. (\ref{eq:u3:xx}) with Lemma~\ref{le:RYI}, we have
\begin{equation}
\begin{aligned}
\label{eq:u3:3}
& \mathbb{E}_{\mathbf{S}} \sqrt{ \frac{1}{T} \sum_{t = 1}^T \sum_{i = \theta + 1}^\infty \left<\mathbf{x}_t, \mathbf{u}_i\right>^2 }\\
\leq & \sqrt{e} \left( \frac{D}{\sqrt{T}} + \sqrt{ \frac{1}{T} \sum_{i = \theta + 1}^\infty \sum_{t = 1}^T  \mathbb{E}_{\mathbf{S}}\left<\mathbf{x}_t, \mathbf{u}_i\right>^2 }\right) \\
=  & D\sqrt{\frac{e}{T}} + \sqrt{\frac{e}{T} \sum_{i = \theta + 1}^\infty \lambda_i}.
\end{aligned}
\end{equation}

To obtain the upper bound of $U_3$, we combine Eqs. (\ref{eq:u3:1}), (\ref{eq:u3:2}) and (\ref{eq:u3:3}).

To sum up, we can complete the proof by
\begin{align*}
\mathcal{R}_{\mathcal{D}}(\mathcal{W})  = & U_1 \cdot U_2 + U_3 \\
\leq & DR \sqrt{\frac{\theta}{T}}  + \frac{R}{T}\sqrt{e \sum_{i = \theta + 1}^\infty \lambda_i} +  \frac{DR\sqrt{e}}{T}\\
\leq & R\sqrt{\frac{1}{T} \left( D^2 \theta +  \sum_{i = \theta + 1}^\infty \frac{e \lambda_i}{T} \right)} +  \frac{DR\sqrt{e}}{T} \\
= & R\sqrt{\frac{1}{T} \left( \sum_{i = 1}^{\theta} D^2 +  \sum_{i = \theta + 1}^\infty \frac{e \lambda_i}{T} \right)} +  \frac{DR\sqrt{e}}{T}.\\
\end{align*}
In the above, the first inequality uses the bounds of $U_1$, $U_2$ and $U_3$ we obtained, the second inequality is owing to the inequality of arithmetic and geometric means for nonnegative numbers: $2 \sqrt{xy} \leq x + y \Leftrightarrow \sqrt{x} + \sqrt{y} \leq \sqrt{2x + 2y}$.
Note that the above bound for $\mathcal{R}_{\mathcal{D}}(\mathcal{W})$ holds for any positive integer $\theta \geq 1$, we obtain the tightest result by minimizing the upper bound, that is
\begin{align*}
\mathcal{R}_{\mathcal{D}}(\mathcal{W}) \leq & R\sqrt{\frac{1}{T}  \min_{\theta \in \mathbb{N}} \left(\sum_{i = 1}^{\theta} D^2 +  \sum_{i = \theta + 1}^\infty \frac{e \lambda_i}{T} \right)} +  \frac{DR\sqrt{e}}{T} \\
=  & R\sqrt{\frac{1}{T}  \sum_{i = 1}^{\infty} \left( D^2 \wedge \frac{e \lambda_i}{T} \right)} +  \frac{DR\sqrt{e}}{T},
\end{align*}
where the equality is owing to that the sequence of eigenvalues $\{ \lambda_i \}$ is in the ascending order.

\end{document}